\documentclass[preprint,review,12pt]{elsarticle}

\usepackage{amssymb}
\usepackage{amsthm}
\usepackage{amsmath}
\usepackage{booktabs}
\usepackage{multirow}
\usepackage{xcolor}

\definecolor{my_color}{named}{black}
\newcommand\cd[1]{\textcolor{my_color}{#1}}
\newcommand\cf[1]{\textcolor{black}{#1}}

\journal{Pattern Recognition}

\begin{document}

\begin{frontmatter}



\begin{titlepage}
\begin{center}
\vspace*{1cm}

\textbf{Graph Perceiver IO: A General Architecture for Graph-Structured Data}

\vspace{1.5cm}

Seyun Bae$^{a}$ (bsu1313@uos.ac.kr), Hoyoon Byun$^b$ (hoyunb@gmail.com), Changdae Oh$^b$ (changdae.oh@uos.ac.kr), Yoon-Sik Cho $^c$ (yoonsik@cau.ac.kr), Kyungwoo Song$^d$$^e$ (kyungwoo.song@gmail.com) \\

\hspace{10pt}
\begin{flushleft}
\small  
$^a$ {Department of Computer Science and Engineering, University of Seoul} \\
$^b$ {Department of Artificial Intelligence, University of Seoul} \\
$^c$ {School of Computer Science and Engineering, Chung-Ang University} \\
$^d$ {Department of Applied Statistics, Yonsei University} \\
$^e$ {Department of Statistics and Data Science, Yonsei University} \\

\vspace{1cm}
\textbf{Corresponding Author:} \\
{\fontsize{8}{8}\selectfont Kyungwoo Song \\
{Department of Applied Statistics, Yonsei University, 50, Yonsei-ro, Seodaemun-gu, Seoul, Republic of Korea} \\
Tel: +82-2-2123-2473 \\
Email: kyungwoo.song@gmail.com\\
\vspace{0.3cm}
Yoon-Sik Cho \\
{School of Computer Science and Engineering, Chung-Ang University, 84, Heukseok-ro, Dongjak-gu, Seoul, Republic of Korea} \\
Email: yoonsik@cau.ac.kr\\}

\end{flushleft}        
\end{center}
\end{titlepage}

\title{Graph Perceiver IO: A General Architecture for Graph-Structured Data}




\author[1]{Seyun Bae}
\author[2]{Hoyoon Byun}
\author[2]{Changdae Oh}
\author[3]{Yoon-Sik Cho\footnote{Corresponding author}}
\author[4,5]{Kyungwoo Song\footnote{Corresponding author; Work partly done at University of Seoul}}
\affiliation[1]{Department of Computer Science and Engineering, University of Seoul}
\affiliation[2]{Department of Artificial Intelligence, University of Seoul}
\affiliation[3]{School of Computer Science and Engineering, Chung-Ang University}
\affiliation[4]{Department of Applied Statistics, Yonsei University}
\affiliation[5]{Department of Statistics and Data Science, Yonsei University}

\begin{abstract}
Multimodal machine learning has been widely studied for the development of general intelligence. Recently, the Perceiver and Perceiver IO, show competitive results for diverse dataset domains and tasks. However, recent works, Perceiver and Perceiver IO, have focused on heterogeneous modalities, including image, text, and there are few research works for graph structured datasets. A graph has an adjacency matrix different from other datasets such as text and image, and it is not trivial to handle the topological information. In this study, we provide a Graph Perceiver IO (GPIO), the Perceiver IO for the graph structured dataset. We keep the main structure of the GPIO as the Perceiver IO because the Perceiver IO already handles the diverse dataset well, except for the graph structured dataset. \cf{The GPIO is a general method that handles diverse datasets, such as graph-structured data, text, and images, by leveraging positional encoding and output query smoothing.} \cf{Compared to graph neural networks (GNNs), GPIO requires lower complexity and can efficiently incorporate global and local information, which is also empirically validated through experiments.} Furthermore, we propose GPIO+ for the multimodal few-shot classification that incorporates both images and graphs simultaneously.
\cf{GPIO achieves higher benchmark accuracy than GNNs across multiple tasks, including graph classification, node classification, and multimodal text classification, while also attaining superior AP and AUC in link prediction. Additionally, GPIO+ outperforms GNNs in multimodal few-shot classification. Our GPIO(+) can serve as a general architecture for handling various modalities and tasks.}
\end{abstract}

\begin{graphicalabstract}
\includegraphics[scale=0.41]{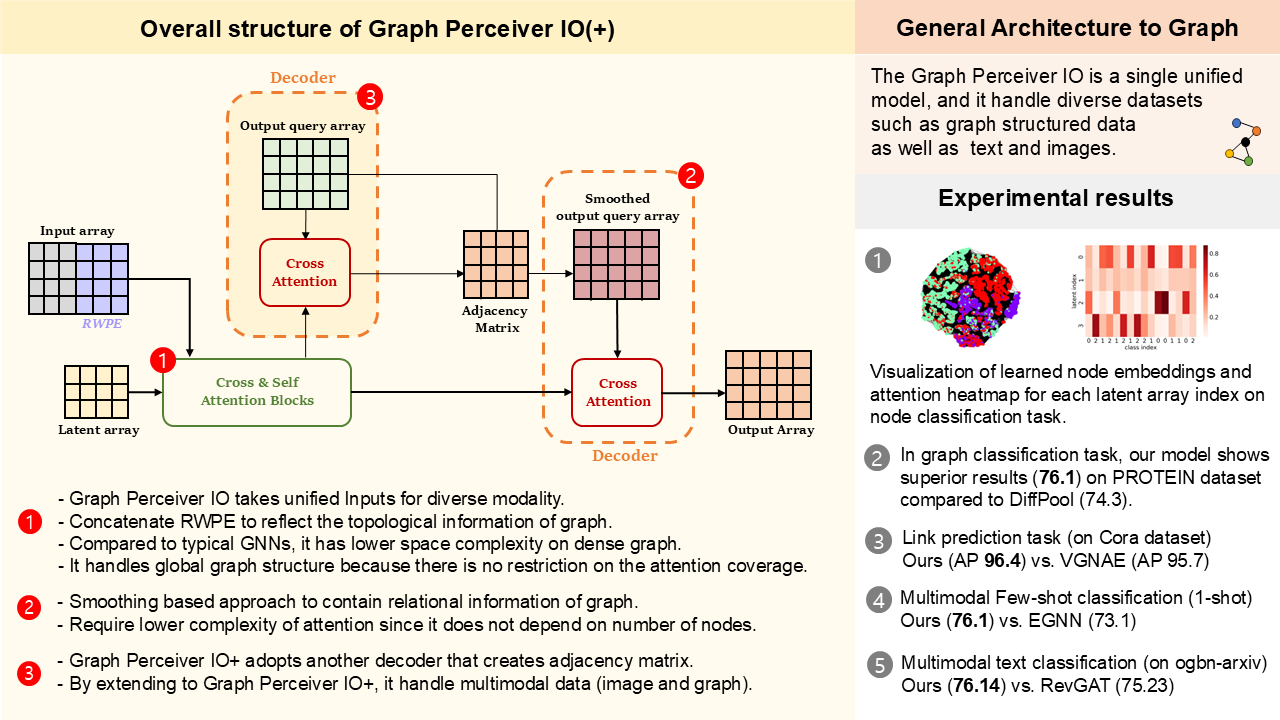}               
\end{graphicalabstract}

\begin{highlights}
\item Graph Perceiver IO generalizes a multimodal method to the graph domain by designing specific graph features to reflect the topological information.
\item Compared to the graph neural networks, Graph Perceiver IO requires a lower complexity, and it can handle the global and local information efficiently. 
\item By extending to two separated decoders, Graph Perceiver IO+ incorporates both images and graphs simultaneously for the multimodal few-shot classification.
\item The Graph Perceiver IO(+) is the single unified model that takes input and output flexibly, and it handles diverse datasets such as graph structured data as well as text and images.

\end{highlights}

\begin{keyword}
Graph Perceiver IO \sep Graph Neural Network  \sep Multimodality
\end{keyword}

\end{frontmatter}


\section{Introduction}
\label{sec:sample1}

Humans who have general intelligence has the capability to handle multiple datasets from different source simultaneously. One of the required building blocks for general intelligence is multimodal learning for a general perception, and there has been much interest in multimodal learning. With the development of the Transformer \cite{vaswani2017attention} and the Vision Transformer (ViT) \cite{dosovitskiy2021an}, there have been multimodal learning approaches that adopts multiple Transformer encoders for each modality \cite{radford2021learning}, and fuses the each information later \cite{li2021align}.
Another remarkable algorithms for multimodal learning are the Perceiver \cite{jaegle2021perceiver} and its extension, the Perceiver IO \cite{jaegle2022perceiver}, general architecture that has structured inputs and outputs. Perceiver IO handles various sizes of inputs and outputs, and they unify the inputs and outputs structures for diverse types of datasets such as images and text. Unified inputs and outputs structures make the single model, Perceiver IO, handle multimodal datasets simultaneously. With Perceiver IO, the specific neural networks for each modality are not required.

Parallel to multimodal learning, there have been many works for the graph structured dataset. A graph has topological information and a graph is known to be a general type that can represent diverse types of the dataset, including image and text \cite{battaglia2018relational}. From its general property, the graph has diverse application areas such as recommender systems \cite{gong2024personalized}, drug discovery \cite{ye2022molecular}, knowledge graph \cite{zhang2023graph} and pedestrian trajectory prediction \cite{zhang2023dual}. The graph has meaningful knowledge information, and the intelligence learned from a graph structured dataset might be important for general intelligence as well as other datasets.
The topological information can be divided by relational information and canonical positional information. Especially, the relational information, described as an adjacency matrix, is a distinct features different from the other dataset. The Perceiver and Perceiver IO flatten all inputs, and they adopt the positional encodings to describe the spatial information or the sequence order. \cf{However, a graph has no such sequence order but explicit relationships, requiring representation distinct from others. Besides, the proper input and output query array for multiple graph-related tasks and multimodal data, including text and images, has not been explored yet.}
%

In this study, we propose the Graph Perceiver IO (GPIO), a general method to handle the graph structured dataset as well as diverse modality. The GPIO takes the raw node features and graph positional encoding to reflect node features and to distinguish various shape of graphs, canonical positional information. Besides, The GPIO can handle relational information of graph through smoothing based output query array design. 
For the GPIO, we propose proper input array and output query array design to enhance the graph task performance while keeping the original model structure of Perceiver IO to preserve the multimodal ablity.
Also, we propose Graph Perceiver IO+ to enhance the multimodal few-shot learning tasks on image and graph data.

The GPIO has three advantages over the traditional GNNs.
First, the GPIO has a lower space complexity. For the given input array, latent array, and output query array, the computation of Perceiver IO does not depend on an adjacency matrix. While the complexity of traditional GNNs is proportional to the squared number of nodes, the GPIO has linear complexity.
Second, the GPIO handles information on both the global and local structure. The GPIO does not have a restriction on the attention areas, and all nodes can interact with attention. It is noted that the locality inductive bias can be injected into the GPIO with the positional encoding and output query array.
Third, the GPIO can handle multimodal data, including the graph, simultaneously. The single unified model, the GPIO, is necessary to consider the multimodal data such as image and graph, while the traditional methods need two networks for image and graph, respectively. \cd{
Our research can contribute to the diagnosis of human diseases. MLP have been studied for diagnosing Alzheimer's Disease (AD) \cite{chelladurai2023fmri} by classifying fMRI images and RNN-LSTM for early diagnosis of Diabetic retinopathy (DR) \cite{ozccelik2023overcoming}. Following this, our multimodal research, including graphs, can be applied to diagnose AD \cite{xu2024interpretable} and DR \cite{yu2023dynamic}.} \cf{For all experiments, here are the URLs of the source code you can reference : https://github.com/MLAI-Yonsei/GPIO} \\

\section{Related Works}

\subsection{Multimodal representation learning}
\cd{
Based on a tremendous amount of image-text pair datasets, contrastive pre-trained multimodal models such as CLIP \cite{radford2021learning}, BLIP-2 \cite{li2023blip} and Alpha-CLIP \cite{sun2024alpha} recently show impressive progress w.r.t generalizability of representation, zero-shot transfer and OOD \cite{mayilvahanan2023does}. 
}
A key factor to their great success is the language-guided visual representation learning scheme. Given a set of image-caption pairs ${(x_{1}, c_{1}), ..., (x_{N}, c_{N})}$, they train an image encoder $E_{img}(.)$ and text encoder $E_{txt}(.)$ such that the similarity $<E_{img}(x_i), E_{txt}(c_i)>$ between positively aligned pair is maximized relative to negatively unaligned pair through InfoNCE \cite{oord2018representation} objective function. In addition to image-text learning, \cd{video-text learning for segmentation \cite{lan2024bidirectional} and video-audio-text sentiment analysis \cite{liu2024sentiment}} and image-speech-text learning \cite{wang2023tetfn} have also studied actively.

\subsection{General architecture}
In another line of work, there are attempts that aim to unify the architecture or learning objective for training in different modalities. Most of these studies mainly utilize the Transformer \cite{vaswani2017attention} architecture which has less data-specific and task-specific inductive bias as the backbone. Instead of relying on modality-specific targets, Data2vec \cite{pmlr-v162-baevski22a} uses the same learning objective for data from different modalities. Based on the combination objective of latent target prediction task \cite{caron2021emerging} and Masked prediction \cite{bao2022beit}, Data2vec shows promising results on three different modalities including text, image and speech. \cite{jaegle2021perceiver} propose a general framework Perceiver to handle arbitrary configurations of different modalities. Without making domain-specific assumptions, Perceiver encodes the data point through iterative self-attention and cross-attention. And they mitigate the quadratic scaling problem of self-attention blocks in Transformer by processing a small set of latent units instead of high-dimensional inputs. 
Recently, Perceiver IO \cite{jaegle2022perceiver} appeared to extend the capable task that the original Perceiver can not perform, and Hierarchical Perceiver \cite{carreira2022hierarchical} improves the performance and efficiency of the Perceiver by introducing locality into the model architecture while preserving its modality-independence property. Our proposed model, GPIO, further extends the accommodatable modality of Perceiver IO to the graph structured dataset.


\subsection{Graph Neural Networks}
The representative method of Graph Neural Networks is GCN \cite{kipf2017semisupervised} which provides spatial domain graph convolution by approximating a spectral graph convolution. GCN propagates a node representation by using a normalized adjacency matrix. APPNP proposes propagation method based on personalized PageRank, which is one of the GCN variants.
\cd{
The memory efficient RevGAT \cite{li2021training} for the deeper and wider models, Dir-GNN \cite{rossi2024edge} which adopts directionality for effective homophily and HiGCN \cite{huang2024higher} for the higher-order interaction have improved classification ability on node-level or graph-level tasks.}
Furthermore, various modified models of GNNs in link prediction tasks ensure reliable performance. \cite{kipf2016variational} predict a link using an inner product between the latent embeddings of target nodes encoded based on a variational autoencoder \cite{Kingma2013AutoEncodingVB}. \cite{ahn2021variational} propose VGNAE that applies $L_2$-normalization before propagation to alleviate the issue that an isolated node embed got close to zero by VGAE \cite{kipf2016variational}. \cd{There are recent works of TokenGT \cite{kim2022pure} that applies the transformer to the graph-structured dataset. In addition, TAPE \cite{he2023harnessing} that utilizes the vast knowledge of LLM to enrich text information and use it as a GNN feature, and GPT4Graph \cite{guo2023gpt4graph} that applies LLM itself to various graph-related tasks.
}

\section{Preliminary}

\subsection{Perceiver}
{\it Jaegle et al.} \cite{jaegle2021perceiver} proposes the Perceiver, a general perception module with iterative attention for handling the diverse and high-dimensional multimodal inputs. The Perceiver has two components of arrays, input arrays $x$, and latent arrays $z$. For given data instances, the Perceiver transforms the given data instances as the inputs arrays
$x \in \mathbb{R}^{M \times C}$ where $M$ and $C$ denote the dataset properties such as the image size and the number of channels. The latent arrays $z \in \mathbb{R}^{N\times D}$ are learnable latent representations, where $N$ and $D$ are the number of latent and the latent dimension, respectively.
The Perceiver adopts the two types of attention, cross-attention and self-attention, and both attentions are query-key-value attention, similar to the Transformer \cite{vaswani2017attention}. First, the cross-attention computes the attention between the latent arrays and the input arrays. The latent arrays are queries, and they attend to the important information from the key, the input arrays. The relevant input features for the given tasks are incorporated into the latent array with cross-attention. Second, the Perceiver adopts the self-attention for the latent arrays.
Attention between latent arrays and input arrays reduces the space and time complexity compared to the self-attention for the input arrays only. The attention complexity of the Perceiver is $O(NM)$ while the complexity of the self-attention for input-image is $O(N^{2})$.
In summary, the Perceiver is composed of the multiple blocks of cross-attention and the latent Transformer that adopts the self-attention for the latent arrays. The input arrays are fed into the Transformer iteratively across the layers. After the multiple blocks, the Perceiver adopts the average pooling over the index dimension and utilizes it to predict the target label. 

\subsection{Perceiver IO}
The Perceiver is a general model that handles diverse datasets with flexible input structures. However, the output structure of the Perceiver is not flexible, and the Perceiver is limited to the classification tasks. To extend the Perceiver into diverse tasks such as language modeling and autoencoding, flexible output structures are necessary. Perceiver IO proposes an additional output query array components $y\in\mathbb{R}^{M \times E}$ for the flexible output structures. For classification task on image, the $M$ denotes the number of images and $E$ denotes the number of classes, respectively. Perceiver IO adds the additional query-key-value attention block, i.e. decoder, on the top of the Perceiver. The additional attention block computes the attention between the output query array and the latent arrays extracted by the Perceiver. The shape of the output query array is a controllable hyper-parameter, and it induces flexible output structures 
that produce any size of outputs.

The Perceiver and Perceiver IO attend to the relationship between the latent arrays and input arrays. They adopt query-key-value attention that is similar to the Transformer. 
Different from the convolutional neural nets (CNN) \cite{he2016deep} or the Vision Transformer (ViT) \cite{touvron2021training}, the Perceiver operates on the individual pixels independently without patches or convolution operations.
However, attention computation is permutation invariant, and the attention-only mechanism is hard to treat the sequence order. To incorporate the order information for sequence and spatial information for images, the Perceiver and the Perceiver IO introduce the learned positional encoding or Fourier-based positional encoding with sinusoid functions.

\subsection{Perceiver IO and Graph Structured Data}
The Perceiver IO is a general architecture, and it has high flexibility for input and output structure.
From its flexibility and generality, The Perceiver and its variants
handle the diverse tasks for image, text, speech, and 3D point clouds. They extend their works to incorporate the heterogeneous multimodal dataset such as audiovisual sequences, and the Perceiver shows competitive results. However, the application of Perceiver is limited to the non-graph structured dataset, and they do not handle the graph structured dataset that has topological information. 

Extending the Perceiver to handle the graph structured dataset is non-trivial. First, the Perceiver is a general architecture, and it already shows the competitive results for diverse tasks. Therefore, simple extensions that do not modify the overall model structures are required. Second, the Perceiver receives the flattened inputs, and incorporating the topological information represented by the adjacency matrix into the Perceiver is challenging. 

\begin{figure*}[t!] 
    \centerline{\includegraphics[width=1\textwidth]{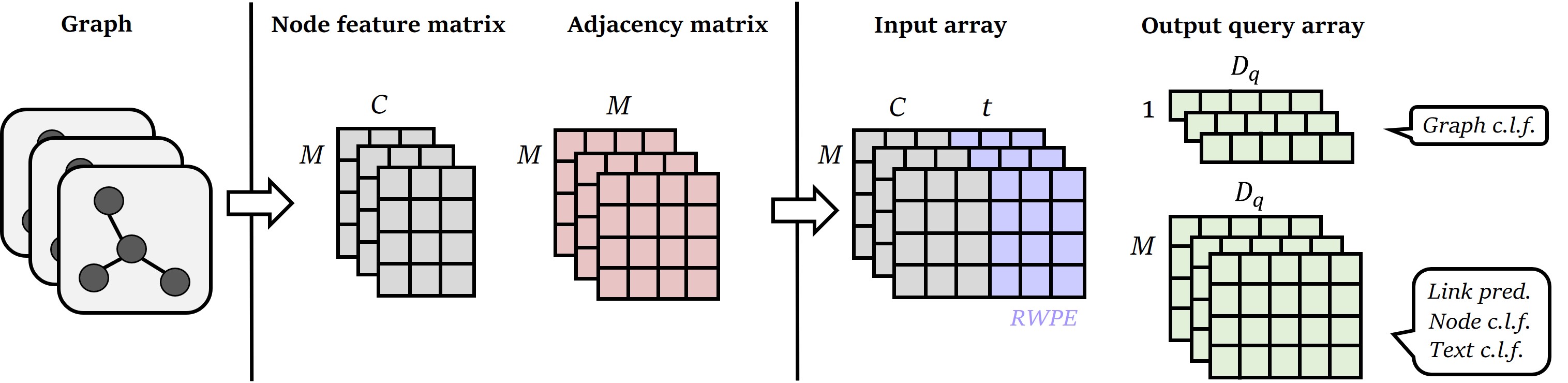}}
    \caption{\cd{A graph-structured dataset splits into a node feature matrix and an adjacency matrix. GPIO construct the input array and the output query array to integrates graph data. Note that the node feature matrix can be text and image feature in multimodal learning.} (i) Input array: Concatenation of node feature and random walk positional embedding ($RWPE$), where $t$ denotes the dimension of $RWPE$. (ii) Output query array: A random initialized $D_q$ dimensional vector used for the graph classification task, and ($M$, $D_q$) for node classification, link prediction and \cd{multimodal text classification.} The output query array can be learnable or fixed, and \textit{c.l.f} denotes the classifier.}
	\label{fig:GPIO_io}
\end{figure*}

\section{Methodology}
\subsection{Overall Structure}
\begin{figure*}[t!] 
    \centerline{\includegraphics[width=0.97\textwidth]{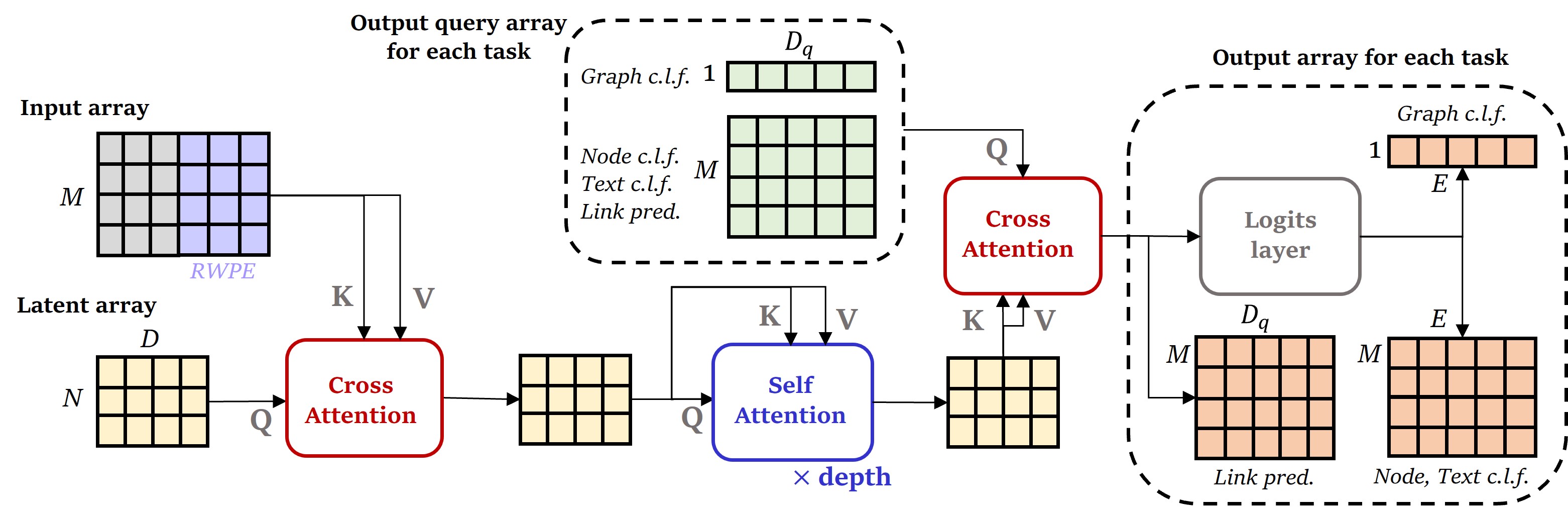}}
    \caption{Overall structure of GPIO. First, the cross-attention between the initial latent and input query enables the latent to absorb the necessary information from the input. Then, the latent progressively encode the salient feature of a given data point through repeated self-attention blocks. Finally, the output query array and final latent communicate via cross-attention to make proper output for each task. 
    $D_{q}$ is an arbitrarily configurable output query array dimension, and $E$ is the number of classes about a given task. 
    The number of depth or layer of self attention block is a controllable hyperparameters.}
	\label{fig:GPIO_framework}
\end{figure*}

Graph data usually consists of $X\in \mathbb{R}^{M \times C}$  matrix, a set of nodes features, and adjacency matrix $A\in \mathbb{R}^{M \times M}$, which represents a set of relational information between nodes. In order to train graphs using Perceiver IO, the input array and the output query array should be constructed by using $X$ and $A$ properly. To handle the graph-structured data, topological information is essential.
We can divide the topological information on the graph as relational information and canonical positional information.
\cd{Both two information is essential for the graph-related tasks requiring the use of adjacency matrix. We propose the methods (i) output query array smoothing to incorporate relational information and (ii) positional embeddings to incorporate canonical positional information. As shown in Figure \ref{fig:GPIO_io}, the methods are implemented by concatenating the node features and positional embeddings and smoothing the node features. } The shape of the output query array depends on the given task, and we will discuss it in Section \textit{Output Query Array}. The overall structure of our proposed model is in Figure \ref{fig:GPIO_framework}.

\subsection{Output Query Array}
\label{sec:oq}

One of the major problems of the GPIO is to incorporate the adjacency matrix. The adjacency matrix that is widely utilized in GNNs is not adaptable for the GPIO. 
Also, the GPIO requires a flexible output structure to handle the diverse graph-related tasks such as link prediction, graph classification, and node classification. For the flexible output array structure, we utilize the output query array of Perceiver IO.
The output query shape can be determined by the task specifically.
It is noted that the output query array can be used as a pretrained embeddings, or it can be learnable parameters.
For graph classification task that requires a single label per graph, we set the output query array size as $1 \times D_{q}$, where $D_{q}$ denotes the output query array dimension.
For node-specific tasks such as node classification and link prediction, we set the output query array shape as $M \times D_q$ where $M$ is the number of nodes. 
The simple choice of the output query array is raw nodes features $X\in \mathbb{R}^{M \times C}$ that is the same as the input array. However, we find that the output query array based on the raw node features tends to show poor performance. Appendix 8.2. presents a detailed result. Furthermore, raw node features do not contain any relational information of adjacency matrix.
Instead of raw node features, we propose a smoothing based output query array that is smoothed features of raw node features $L$ times \cite{wu2019simplifying}. The following Eq. \ref{eq:smoothing} denotes the $L$ times smoothed node features, the output query array of the GPIO,
\begin{equation}
\widehat{X}=\left ( \widetilde{D}^{-\frac{1}{2}}\widetilde{A}\widetilde{D}^{-\frac{1}{2}} \right )^{L}X \label{eq:smoothing}
\end{equation}
where $\widetilde{A} = A + I$ denotes a adjacency matrix with self-loop and $\widetilde{D}_{ii} = \sum_{j=0}\widetilde{A}_{ij}$ denotes node degree matrix of $\widetilde{A}$.
SGC \cite{wu2019simplifying} removes nonlinear function between GCN layers and collapse repeated multiplication of matrix into $L$-th power of matrix like Eq. \ref{eq:smoothing} which extracts smoothed features of $X$. As a result, extracted features $\widehat{X}$ tends to have similar representation to nearby nodes and help to make similar predictions. That is, the SGC incorporates adjacency matrix into its input to get $L$-hops neighbor information. In a similar way, we utilize adjacency matrix to obtain $\widehat{X}$ as an output query array which contain relational information.

Besides, we adopt the APPNP smoothing \cite{gasteiger2018combining} as an output query array as shown in Eq. \ref{eq:appnp_smoothing}. 
\begin{eqnarray}
X^{(0)} &=& X,\nonumber \\
X^{(l)} &=& (1-\alpha)\widetilde{D}^{-\frac{1}{2}}\widetilde{A}\widetilde{D}^{-\frac{1}{2}}X^{(l-1)}+\alpha X^{(0)}, \label{eq:appnp_smoothing} \nonumber \\
\widehat{X}_{APPNP} &=& X^{(L)}
\end{eqnarray}
APPNP adopts personalized propagation, and it provides more data specific smoothing methods. We empirically find that the output query array with a smoothing method is one of the key factors for the GPIO. Section 4.5. presents complexity analysis of output query array.

\subsection{Input Array}
\label{sec:iq}
The GPIO requires the flexible input structure as the Perceiver or Perceiver IO.
For image and text dataset, {\it jaegle et al.} propose Fourier embedding to consider the spatial or sequential information, but it shows poor performance for graph classification \cite{jaegle2022perceiver}. Appendix 8.2.3. presents a detailed result. To reflect the topological information, specifically canonical positional information, with the original Perceiver input array structure, we adopt the random walk positional embedding ($RWPE$) \cite{dwivedi2022graph}.
Random walk operators are defined as $R = AD^{-1}$ where $A \in \mathbb{R}^{M \times M}$ is the adjacency matrix, and $D$ is the degree matrix. $RWPE$ of node $i$ is defined as
\begin{equation}
    RWPE_{i} = [R_{ii}, R_{ii}^{2}, ... ,R_{ii}^{t} ]  \in \mathbb{R}^{t},
\end{equation}
where $R_{ii}^{t}$ is the probability of starting from the $i$ node, walking $t$ times, and returning to the $i$ node.
{\it dwivedi et al.} adopt $RWPE$ as positional encoding initialization for distinguishing several cases of non-isomorphic graphs and structurally different nodes which messeage-passing GNNs and 1-WL \cd{\cite{weisfeiler1968reduction}} fail to distinguish.
The choice of sufficiently large $t$ gives a unique node representation to distinguish each node and graph.
Besides the node features, we concatenate the $RWPE$ to obtain nodes representation distinguishing non-isomorphic graphs and structurally different nodes. There is no restriction on the attention coverage, and it makes the GPIO handles the global structure information efficiently. It is noted that the injection of $RWPE$ into the GPIO can be interpreted as an inductive bias to consider the local information as well as the global information. 
Compared to typical GNNs, the Perceiver has lower space complexity in terms of the input array. GNNs requires $O(M \times C + M \times M)$ on dense graph, while the GPIO needs $O(M(C+t))$, where $t$ is the controllable hyper-parameters. \cf{Consequently, GPIO is not proportional to the number of edges, making it more efficient on dense graphs. Experiments on Barabási-Albert (BA) graphs of varying densities confirm that GNN memory usage grows with density, whereas GPIO depends only on the number of nodes $M$. This highlights GPIO’s space efficiency for large, dense graphs. The specific experimental analysis is in the Appendix 8.5.2.}

\subsection{Model Analysis}
\cd{
We demonstrate how the GPIO leverages graph structures via adjacency matrix $A$. The decoder in the GPIO computes the attention scores between latent array $z$ and output query array $X$. The attention score matrix $\mathbf{S}\in \mathbb{R}^{M \times N}$ can be represented as follows, where $W_{Q}$ and $W_{K}$ denote query and key projection matrix with hidden size $d$.}
\begin{equation}
\mathbf{S} = \frac{(XW_{Q})(zW_{K})^\top}{\sqrt{d}} \label{eq:QKattn}
\end{equation}
\cd{
Each element in $\mathbf{S}$ is a score indicating how much each latent is focusing on which node features. This means that $z$ attends globally, regardless of whether the node feature is connected to any of its neighbors. By taking feature smoothing $\widehat{X}$, the GPIO exploits $A$ to propagate the global score to its local neighbors. The propagated score matrix $\mathbf{\widehat{S}}$  can be derived as follows, where $\Ddot{A}$ = $\widetilde{D}^{-\frac{1}{2}}\widetilde{A}\widetilde{D}^{-\frac{1}{2}}$.}
\begin{align}
\mathbf{\widehat{S}} = \frac{(\widehat{X}W_{Q})(zW_{K})^\top}{\sqrt{d}} 
&= \frac{(\Ddot{A}^{L}XW_{Q})(zW_{K})^\top}{\sqrt{d}} \nonumber \\
&= \Ddot{A}^{L} \mathbf{S}, \label{eq:QKattnSmooth}
\end{align}
\cd{
The relational information $\Ddot{A}$ is responsible for propagating the weighted score to its neighbor that are $L$-hops away. The global focus of a latent spreads its impact across local neighbor nodes, ensuring that nodes at least $L$-hops away have similar scores. Detailed experimental results are covered in \textit{Discussion}.}

\cd{
An alternative is APPNP smoothing, which prevents oversmoothing compared to Eq. \ref{eq:QKattnSmooth} to propagate to more distant neighbors.}
\begin{equation}
\mathbf{\widehat{S}_{\mathit{APPNP}}}=(\alpha \sum_{l=0}^{L-1}(1-\alpha)^{l}\Ddot{A} + (1-\alpha)^{L}\Ddot{A}^{L})\mathbf{S}
\end{equation}
\cd{The global attention score that a latent assigns to a particular node is propagated to nodes within an $L$-hops with different weights based on $\Ddot{A}$, allowing the GPIO to handle graph structures.}

\subsection{Complexity Analysis}

The GPIO requires a single adjacency matrix-related operation to make the smoothed output query array before the model training. The smoothing operation is only once conducted, and it never appears during the model training. The mechanism is similar to the Simplifying Graph Convolutional (SGC) \cite{wu2019simplifying} that only performs single adjacency matrix-related operation. Therefore, the GPIO has an efficient computation time compared to the other GNNs.
Besides, there is no attention between inputs, but there is attention between the latent array and the input array, or the output query array and the input array. 
The complexity of attention between latent arrays is independent of the number of nodes, and the complexity of attention between latent and input arrays depends on the number of nodes linearly. However, the current Transformer variants, such as Graph Transformer architecture \cite{dwivedi2020generalization} adopt the input-wise attention, and the computation time is proportional to the squared number of nodes. Therefore, GPIO has advantages in the time complexity as well as its simplicity and multimodal learning.


%

\subsection{Graph Perceiver IO+}
\begin{figure*}[h!] 
    \centerline{\includegraphics[width=1.0\textwidth]{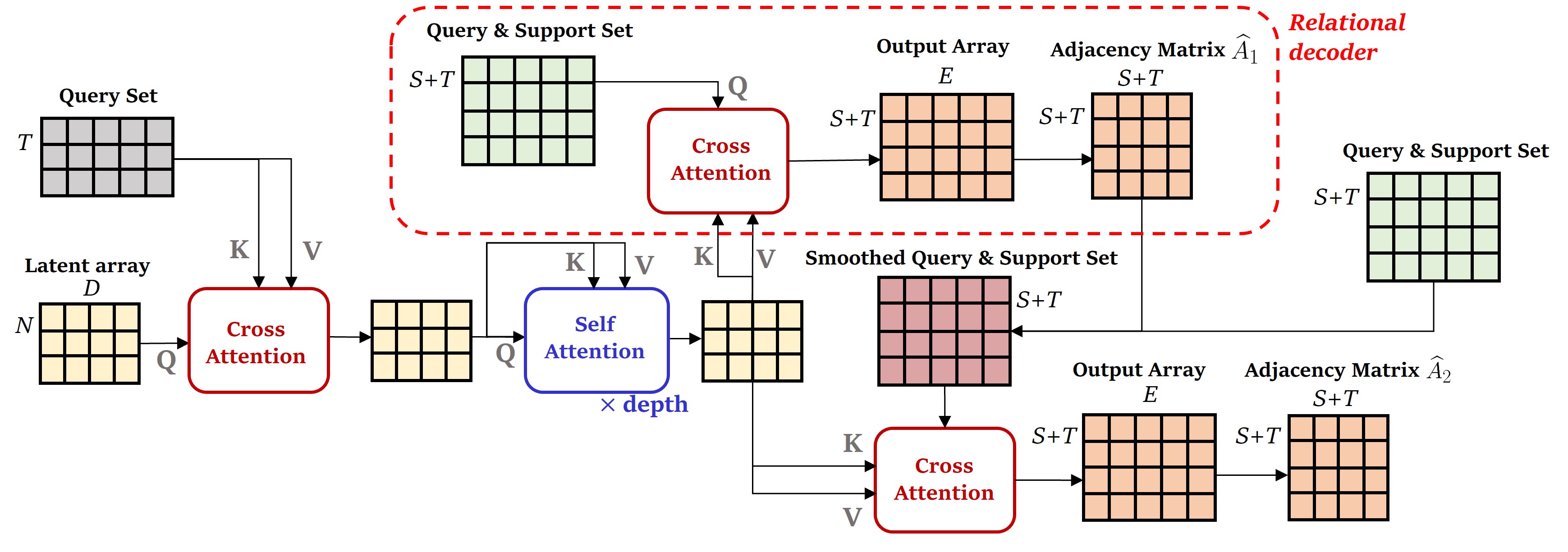}}
    \caption{Overall structure of GPIO+. \cd{The red dashed area is the relational decoder introduced for the extension to GPIO+. $\widehat{A}_{1}$ gives rich information to classify $\mathcal{Q}$. The second decoder performs multimodal few-shot image classification.}}
	\label{fig:GPIO+_framework}
\end{figure*}

\cd{The few-shot classification task is to classify query set $\mathcal{Q}$ which contains unlabeled $T$ samples, when support set $\mathcal{S}$ which contains labeled $S$ samples is given. $n$-way $k$-shot classification problem means that support set $S$ contains number of $n$ classes and each class has number of $k$ labeled samples. 
Constructing graph-structures between $\mathcal{Q}$ and $\mathcal{S}$ provides rich relational information to classify which class a sample from $\mathcal{Q}$ belongs to among the $\mathcal{S}$. 
To enhance the multimodal few-shot task performance, we additionally propose Graph Perceiver IO+ (GPIO+), that handle both image features and relational information of each set by adopting \textit{Relational decoder}. As seen in Figure \ref{fig:GPIO+_framework}, the \textit{Relational decoder} is extension of the original GPIO's decoder that decodes relational information between $\mathcal{Q}$ and $\mathcal{S}$ \cite{kim2019edge}. 
We configure input array as samples of $\mathcal{Q}$ and output query array as samples of $\mathcal{Q}$ and $\mathcal{S}$ in the both decoders. The \textit{Relational decoder} creates adjacency matrix $\widehat{A}_{1} \in \mathbb{R}^{(S+T) \times (S+T)}$, and the second decoder that is original decoder of GPIO creates $\widehat{A}_{2} \in \mathbb{R}^{(S+T) \times (S+T)}$. The \textit{Relational decoder} takes the unsmoothed samples of $\mathcal{Q}$ and $\mathcal{S}$ as an output query array, and creates adjacency matrix $\widehat{A}_{1}$ which is obtained by inner product of output array itself. $\widehat{A}_{1}$ represents relational information of each pair of $\mathcal{Q}$ and $\mathcal{S}$.
The second decoder takes a smoothed output query array which is smoothed by $\widehat{A}_{1}$, and it generates an adjacency matrix $\widehat{A}_{2}$. $\widehat{A}_{2}$ is final output to classify $\mathcal{Q}$. The edge label matrix $Y_e \in \mathbb{R}^{(S+T) \times (S+T)}$ with a value of 1 if the class is the same and 0 if the class is different for all query set and support set pairs.
GPIO+ follows episodic training, which mimics few-shot learning setting of test time, and the each label in train and test time is sampled from mutually exclusive set of all classes.}
In each episodic, the final loss $\mathcal{L}$ is obtained by computing loss between $Y_{e}$ and weighted average of $\widehat{A}_{1}$ and $\widehat{A}_{2}$ such that \begin{equation}
\widehat{A} = \beta\widehat{A}_{1} +  (1-\beta)\widehat{A}_{2}, \text{ and }
\mathcal{L}= \mathcal{L}_{bce}(Y_{e}, \widehat{A})
\end{equation}
where $\beta$ is controllable hyperparameter and $\mathcal{L}_{bce}$ is binary cross entropy loss. The final class of query set selects one of the class of all support set with the highest edge predicted value of $\widehat{A}$ \cite{kim2019edge}.

\section{Results}
To validate our proposed model, the GPIO, we conduct comprehensive experiments, node classification, graph classification, link prediction, multimodal few-shot image classification, and multimodal text classification. Appendix 8.1. provides the more detailed experimental settings.

\subsection{Link Prediction}
\begin{table*}[h!]
\centering
\resizebox{1\textwidth}{!}{%
\begin{tabular}{lcccccc}
\hline
\multicolumn{1}{c}{\multirow{2}{*}{Model}} & \multicolumn{2}{c}{Cora}                            & \multicolumn{2}{c}{CiteSeer}                        & \multicolumn{2}{c}{PubMed}                         \\
\multicolumn{1}{c}{}                       & AUC                      & AP                       & AUC                      & AP                       & AUC                      & AP                      \\ \hline
Spectral Clustering*                       & 84.6$\pm{0.01}$          & 88.5$\pm{0.0}$              & 80.5$\pm{0.01}$              & 85.0$\pm{0.01}$              & 84.2$\pm{0.02}$              & 87.8$\pm{0.01}$          \\
DeepWalk*                                  & 83.1$\pm{0.01}$          & 85.0$\pm{0.0}$              & 80.5$\pm{0.02}$              & 83.6$\pm{0.01}$              & 84.4$\pm{0.0}$              & 84.1$\pm{0.0}$          \\
DGI*                                       & 89.8$\pm{0.8}$           & 89.7$\pm{1.0}$              & 95.5$\pm{1.0}$              & 95.7$\pm{1.0}$              & 91.2$\pm{0.6}$              & 92.2$\pm{0.5}$              \\
ARGVA*                                     & 92.4$\pm{0.004}$         & 93.2$\pm{0.003}$              & 92.4$\pm{0.003}$              & 93.0$\pm{0.003}$              & 96.8$\pm{0.001}$              & 97.1$\pm{0.001}$               \\
GIC \cite{DBLP:conf/pakdd/MavromatisK21}                                        & 90.0$\pm{1.0}$           & 89.9$\pm{1.3}$              & 95.8$\pm{0.7}$              & 95.8$\pm{0.9}$              & 90.9$\pm{1.0}$              & 91.6$\pm{0.9}$                     \\
VGAE \cite{kipf2016variational}                                       & 95.2$\pm{0.5}$           & 94.7$\pm{0.6}$              & 92.0$\pm{1.7}$                & 91.6$\pm{1.7}$              & 95.6$\pm{0.7}$              & 95.3$\pm{0.6}$                     \\
GNAE \cite{ahn2021variational}                                       & 95.6$\pm{0.7}$           & \underline{96.0}$\pm{0.8}$  & \textbf{97.2}$\pm{0.5}$  & \textbf{97.3}$\pm{0.4}$  & \textbf{97.7}$\pm{0.2}$  & \underline{97.6}$\pm{0.2}$    \\
VGNAE \cite{ahn2021variational}                                     & \underline{95.8}$\pm{0.6}$  & 95.7$\pm{0.8}$           & 96.8$\pm{0.6}$           & 96.7$\pm{0.6}$           & 97.3$\pm{0.1}$           & 97.2$\pm{0.2}$             \\ \hline
GPIO                          & \textbf{95.9}$\pm{0.3}$              & \textbf{96.4}$\pm{0.2}$             & \underline{96.8}$\pm{0.3}$              & \underline{97.2}$\pm{0.2}$              & \underline{97.6}$\pm{0.08}$                & \textbf{97.8}$\pm{0.07}$                                                      \\ \hline
\end{tabular}
}
\caption{Performance of link predictions. GPIO shows the competitive results with the well-known link prediction model VGAE, which adopts the graph convolutional neural networks. * denotes the reported performance in \cite{DBLP:conf/pakdd/MavromatisK21}.}
\label{tab:link_prediction_table}
\end{table*}

Besides the classification task, we validate our models on the link prediction task. Table \ref{tab:link_prediction_table} denotes the results of the link prediction
task. 
We utilize the features extracted by the cross attention between the output query array and latent array for the link prediction task. The GPIO shows the competitive performance across all dataset. We provide detailed experimental settings and t-SNE plot to compare the performance of the three models in Appendix 8.2.1.
%
%
%

\subsection{Node Classification}

\begin{table*}[h!]

\centering
\resizebox{0.8\textwidth}{!}{%
\begin{tabular}{lcccccc}
\hline
\multicolumn{1}{c}{\multirow{2}{*}{Model}} & \multicolumn{2}{c}{Cora}      & \multicolumn{2}{c}{CiteSeer}  & \multicolumn{2}{c}{PubMed}    \\
\multicolumn{1}{c}{}                       & Fixed         & Random        & Fixed         & Random        & Fixed         & Random        \\ \hline
GCN                                        & 81.4$\pm{0.7}$          & 78.7$\pm{1.7}$          & 71.1$\pm{0.7}$          & 68.1$\pm{1.7}$          & 78.9$\pm{0.6}$          & 77.3$\pm{2.4}$          \\
GAT                                       & 83.0$\pm{0.5}$          & 80.9$\pm{1.5}$          & 70.9$\pm{0.5}$          & \underline{68.8}$\pm{1.7}$          & 78.9$\pm{0.4}$         & 77.6$\pm{2.4}$          \\
Cheb                                       & 80.5$\pm{1.0}$          & 77.0$\pm{2.7}$          & 69.8$\pm{1.2}$          & 67.2$\pm{2.1}$          & 78.2$\pm{0.6}$          & 75.4$\pm{2.5}$          \\
SGC                                      & 81.7$\pm{0.1}$          & 80.1$\pm{1.9}$          & \underline{71.3}$\pm{0.2}$          & 68.5$\pm{2.0}$          & 78.9$\pm{0.0}$          & 76.6$\pm{2.4}$          \\
ARMA                                      & 82.2$\pm{0.9}$          & 79.8$\pm{1.7}$          & 71.0$\pm{0.6}$          & 67.9$\pm{1.9}$          & 78.8$\pm{0.3}$          & 77.6$\pm{2.2}$          \\
APPNP                                      & \underline{83.3}$\pm{0.5}$ & \textbf{82.1$\pm{1.5}$} & \textbf{71.7$\pm{0.5}$} & \textbf{69.8$\pm{1.8}$} & \textbf{80.1$\pm{0.2}$} & \underline{79.1}$\pm{2.3}$          \\ \hline
GPIO                          & \textbf{83.9$\pm{0.6}$}          & \underline{81.6}$\pm{1.8}$          & 70.1$\pm{1.0}$          & 68.1$\pm{1.7}$          & \underline{79.9}$\pm{0.4}$          & \textbf{79.6$\pm{2.2}$} \\ \hline
\end{tabular}
}
\caption{Accuracy for the node classification. The GPIO shows the competitive results on PubMed. Fixed and Random denotes the dataset split methods from \cite{Fey:2019wv}.}
\label{tab:noderesults}
\end{table*}

\begin{figure}[h!]
    \centering
    \includegraphics[width=.3\columnwidth]{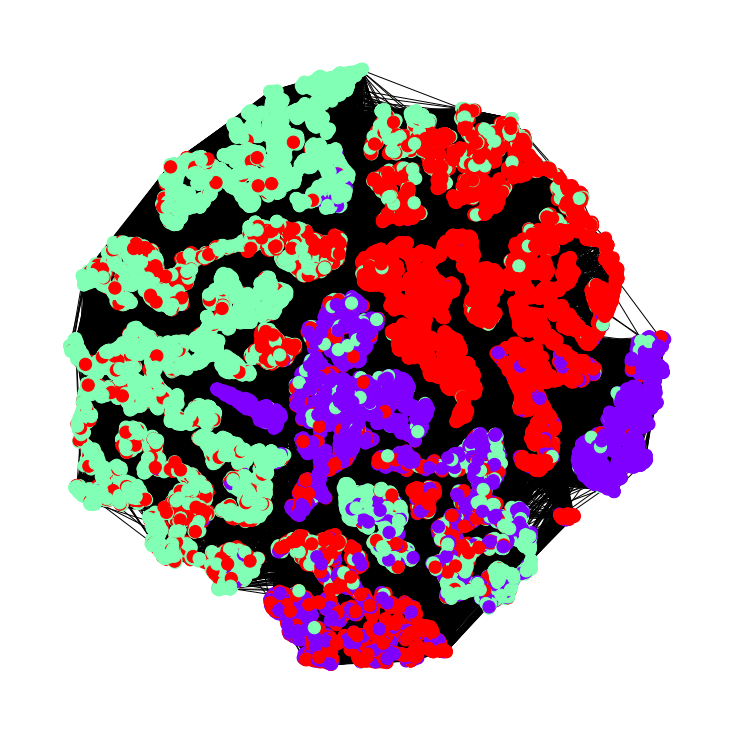}
    \includegraphics[width=.3\columnwidth]{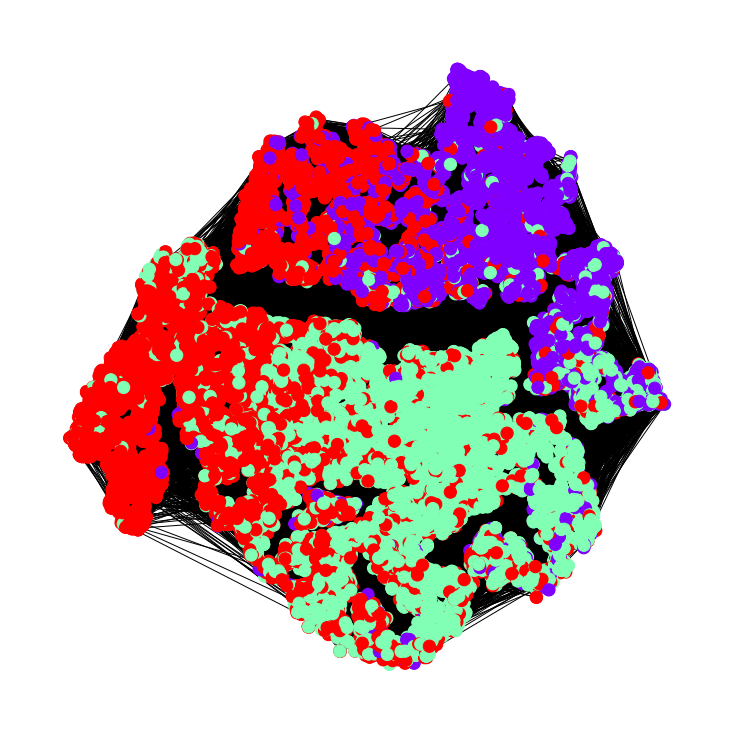}
    \includegraphics[width=.3\columnwidth]{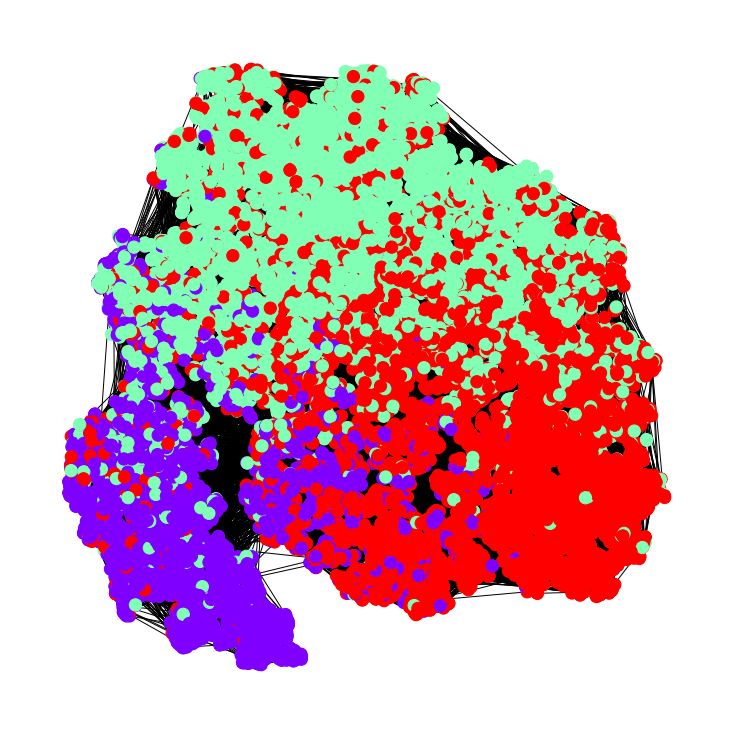}
\caption{
t-SNE \cite{van2008visualizing} visualization of learned nodes embedded by the GPIO (left), APPNP (middle), GAT (right) on PubMed dataset. The learned embedding of GPIO shows a relatively large uniformity compared to the APPNP and GAT. Large uniformity denotes that a feature distribution utilizes maximal information, and it has a positive correlation with the downstream task performance \cite{wang2020understanding}.
}
\label{fig:tsne_node_pubmed}
\end{figure}

For node classification experiments, we adopt the three benchmark data, Cora \cite{cora}, CiteSeer \cite{citeseer}, and PubMed \cite{pubmed}. We follow all experimental settings and GNN baseline from \cite{Fey:2019wv}, and we repeat the experiment 100 times. Table \ref{tab:noderesults} denotes the accuracy of node classification task. The GPIO shows the competitive results on the Cora and PubMed dataset.

\cf{Besides, for the PubMed dataset, which has 3 classes and 500-dimensional features, we extracted high-dimensional embeddings from the penultimate layer of each model and visualized their 2D projection in Figure \ref{fig:tsne_node_pubmed}. Each color represents a class. The embedding visualization confirms the GPIO has a relatively high uniformity embedding space.}

\subsection{Graph Classification}

\begin{table*}[h!]
\centering
\resizebox{0.8\textwidth}{!}{%
\begin{tabular}{lccccc}
\hline
\multicolumn{1}{c}{Model} & MUTAG         & PROTEINS      & \begin{tabular}[c]{@{}c@{}}IMDB-\\ BINARY\end{tabular} & \begin{tabular}[c]{@{}c@{}}REDDIT-\\ BINARY\end{tabular} & COLLAB        \\ \hline
GCNWithJK                 & 72.9$\pm{12.0}$          & 72.6$\pm{3.6}$          & 73.2$\pm{5.0}$                                                   & 89.4$\pm{2.9}$                                                     & \textbf{81.5}$\pm{2.1}$ \\
SAGPool                   & 74.0$\pm{8.7}$          & 72.3$\pm{2.8}$          & 72.3$\pm{4.7}$                                                   & 89.0$\pm{2.1}$                                                     & 78.9$\pm{1.0}$          \\
DiffPool                  & \underline{84.6}$\pm{8.7}$          & \underline{74.3}$\pm{6.4}$          & \textbf{74.8}$\pm{4.8}$                                          & \textbf{92.7}$\pm{2.0}$                                            & 79.4$\pm{1.9}$          \\
GCN                       & 70.7$\pm{11.0}$          & 72.2$\pm{2.4}$          & 74.2$\pm{4.4}$                                                   & 89.1$\pm{2.0}$                                                     & \underline{81.0}$\pm{1.4}$          \\
GraphSAGE                 & 75.1$\pm{11.4}$          & 74.1$\pm{2.3}$          & 73.2$\pm{4.4}$                                                   & 90.7$\pm{2.3}$                                                     & 79.8$\pm{1.1}$          \\
GIN0                      & 81.9$\pm{8.0}$          & 73.1$\pm{3.8}$          & \underline{73.7}$\pm{4.1}$                                                   & 90.9$\pm{2.1}$                                                     & 80.5$\pm{1.9}$          \\
Set2SetNet                & 74.5$\pm{11.9}$          & 74.1$\pm{3.8}$          & 72.7$\pm{5.0}$                                                   & 90.3$\pm{2.4}$                                                     & 79.4$\pm{1.7}$          \\
SortPool                  & 83.0$\pm{9.0}$          & 73.9$\pm{4.5}$          & 71.8$\pm{3.0}$                                                   & 84.3$\pm{5.0}$                                                     & 77.8$\pm{1.6}$          \\
ASAP                      & 78.7$\pm{11.8}$          & 74.0$\pm{3.0}$            & 72.2$\pm{4.3}$                                                   & OOM                                                      & 79.4$\pm{1.7}$          \\ \hline
GPIO         & \textbf{86.1}$\pm{6.9}$ & \textbf{76.1}$\pm{3.0}$ & 72.9$\pm{4.4}$                                                   & \underline{91.6}$\pm{2.3}$                                                     & 79.1$\pm{2.5}$          \\ \hline
\end{tabular}
}
\caption{Accuracy of the graph classification tasks. GPIO shows the superior performance on MUTAG and PROTEINS dataset that has node features. The topological information is significant because IMDB, REDDIT, and COLLAB have no node features. GPIO shows the competitive results even though there are no node features.}
\label{tab:graph_classification}
\end{table*}

We validate our models on the graph classification task. Graph classification tasks require the understanding of both the global and local graph structures. We repeat the experiments ten times, and we follow the same experimental protocols with the \cite{Fey:2019wv}. Table \ref{tab:graph_classification} denotes the accuracy of the graph classification on diverse benchmark dataset. For MUTAG \cite{debnath1991structure} and PROTEINS \cite{debnath1991structure}, GPIO shows superior performance compared to the recent GNNs. In short, GPIO has a high capacity to incorporate both the node features and link information simultaneously. 
It is noted that MUTAG and PROTEINS have node features while IMDB, REDDIT, and COLLAB \cite{yanardag2015deep} have no node features. 
Even though there are no node features, the GPIO has competitive results with the recent GNNs. It denotes that the GPIO handles the topological information of graph-structured data with the positional embeddings.

\subsection{Multimodal Few-shot classification}

\begin{table}[!h]
\centering
\resizebox{0.5\textwidth}{!}{%
\begin{tabular}{lccc}
\hline
Model             & 1-shot & 3-shot & 5-shot \\ \hline
MAML+Transduction  & 50.8*  & -      & 66.2** \\
TPN              & 53.8*  & -      & 69.4** \\
EGNN (original)    & 59.2   & 71.2   & 75.7   \\
EGNN†       & \underline{73.1}   & \underline{83.5}   & \textbf{87.9}   \\
GPIO+†     & \textbf{76.1}   & \textbf{83.7}   & \underline{86.9}   \\ \hline
\end{tabular}%
}
\caption{The 5-way, 1-shot, 3-shot and 5-shot few-shot classification accuracies on
miniImageNet dataset. * and ** denotes the reported performance in \cite{liu2018learning} and \cite{kim2019edge}, respectively. † denotes result to use pre-trained feature extractor.}
\label{tab:fs-table}
\end{table}

We perform a few-shot image classification task to evaluate GPIO+ for image-graph multimodal learning. We follow public source code of \cite{kim2019edge} as an experimental settings. Also, we employ pretrained VGG16 as feature extractor, and change the number of training and test iterations to 10K and 1K, respectively. Table \ref{tab:fs-table} shows the results on 5-way setting on miniImageNet dataset. GPIO+ shows superior performance, compared to EGNN in 1-shot and 3-shot settings.
The \textit{relational decoder} of GPIO+ takes raw image features as the output query array to generate the relational information of the images (nodes). The graph structure of images (nodes) is generated in this step. The second decoder takes smoothed image features as output query array to incorporate graph structure of nodes. Using two separated decoders, GPIO+ handle multimodal data such as image and graph.

\subsection{\cd{Multimodal Text Classification}}

\begin{table}[b!]
\centering
\resizebox{0.5\textwidth}{!}{%
\begin{tabular}{lcc}
\hline
Model  & obgn-products & ogbn-arxiv \\ \hline
MLP    & 0.7429$\pm{0.0075}$        & 0.7265$\pm{0.0014}$     \\
GCN    & 0.7561$\pm{0.0202}$        & 0.7409$\pm{0.0035}$     \\
SAGE   & 0.7428$\pm{0.0115}$        & 0.7355$\pm{0.0037}$     \\
RevGAT & 0.7663$\pm{0.0064}$        & 0.7523$\pm{0.0025}$     \\
GPIO   & \textbf{0.7763$\pm{\textbf{0.0007}}$}        & \textbf{0.7614$\pm{\textbf{0.0006}}$}     \\ \hline
\end{tabular}
}%
\caption{\cd{Text classification accuracy for the ogbn-products and ogbn-arxiv datasets \cite{hu2020open}. For ogbn-products, we used the same subset data as in \cite{he2023harnessing}.}}
\label{tab:text_classification}
\end{table}

\cd{We conduct evaluation on OGB datasets \cite{hu2020open} which contains large graphs for text-graph multimodal learning. We follow \cite{he2023harnessing} as GNNs baseline and experimental settings and employ fine-tuned language model as text feature extractor which is size 129M \cite{he2023harnessing}. The nodes of the graph consist of the title, summary, or product description of the article, which are composed of the citation network or the concurrent purchase network, respectively. The model should classify these text nodes into the correct class using topological information. Dataset statistics and are in the Appendix 8.4. In table \ref{tab:text_classification}, GPIO outperforms GNNs on all datasets and handles large structured graph datasets effectively. The result illustrates that our model performs well in text classification by utilizing the text and its topological information. }


\section{\cd{Discussion}}

 \begin{figure*}[h!] 
    \centerline{\includegraphics[width=1.0\textwidth]{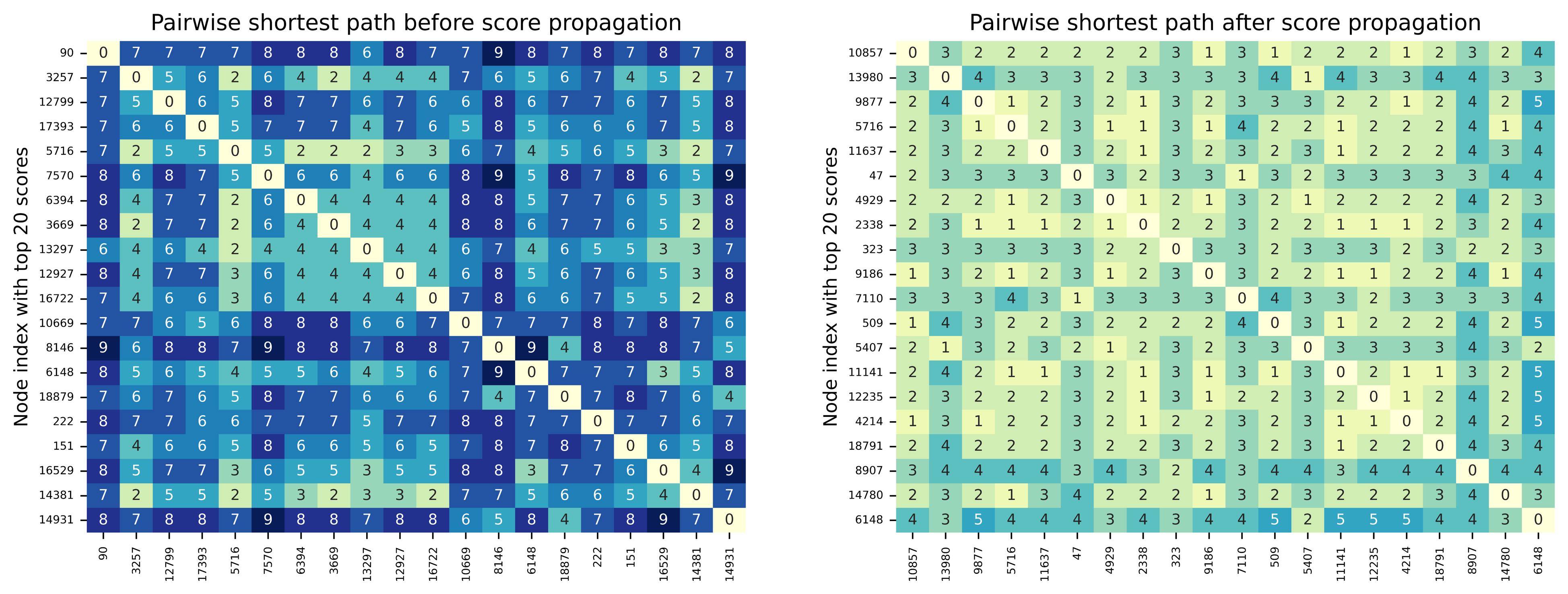}}
    \caption{The pairwise shortest path heatmap on PubMed \cite{pubmed} dataset. The left side represents before 2-hops away score propagation and the right side represents after.}
	\label{fig:distance_heatmap}
\end{figure*}

\cd{
We explore how GPIO propagates global information locally with feature smoothing through a visualization. We take one latent that has absorbed the required global information from the input and compute the attention scores $\mathbf{s}\in \mathbb{R}^{M \times 1}$ with the output query array to select the top 20 nodes with the highest attention score out of $M$ nodes. That is, these are the nodes that the latent has chosen to focus on. We computed the pairwise shortest path of these nodes to see their distance. In Figure \ref{fig:distance_heatmap}, the nodes are relatively farther apart before propagation than after. Feature smoothing has changed the global focus to a neighboring local focus. Also, although we used $\widehat{X}$ with $L=2$, some nodes focused by latent still have distances greater than 2. Therefore, GPIO can incorporate the global and local graph structures efficiently.}

\cd{
In graph-related tasks, understanding complex graph structures requires the ability to distinguish between two graphs with different structures. The classic way to tell if two graphs are isomorphic is the 1-WL test \cite{weisfeiler1968reduction}: if two graphs are not isomorphic using the 1-WL test, they have different structures. However, the 1-WL test does not guarantee isomorphism. In other words, there are cases where the 1-WL test fails to distinguish between graphs. The typical messege-passing GNNs have an expressive power of up to 1-WL. The key to competing with GNN is to be as expressive as or more expressive than 1-WL. A simple choice would be to sequentially give each node a position embedding, like Perciever IO, but that would be less expressive, as the only thing that distinguishes the graph is the number of nodes. }
\begin{figure}[t!]
    \centering
    \includegraphics[width=.32\columnwidth]{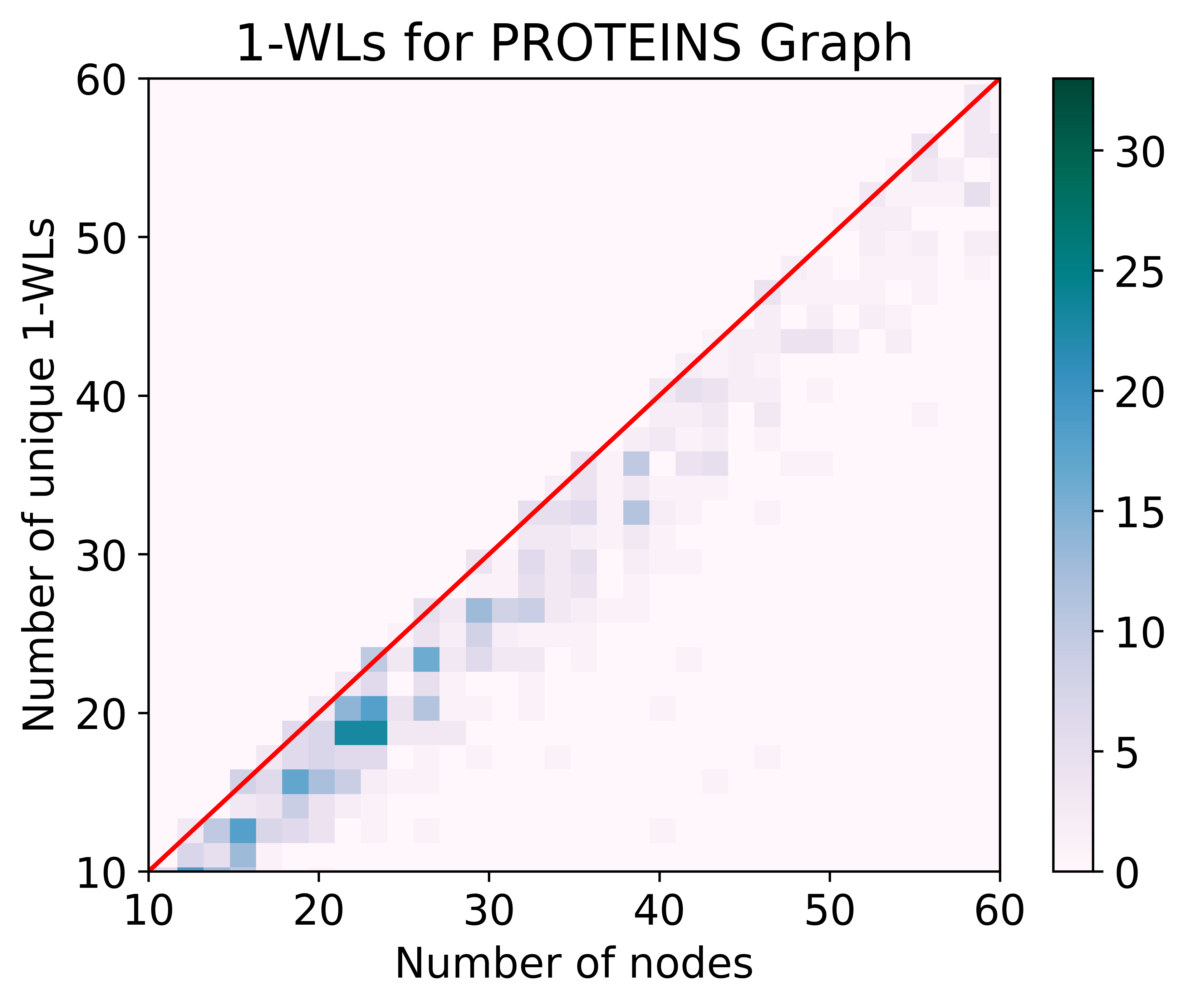}
    \label{fig:protein_wl}
    \includegraphics[width=.32\columnwidth]{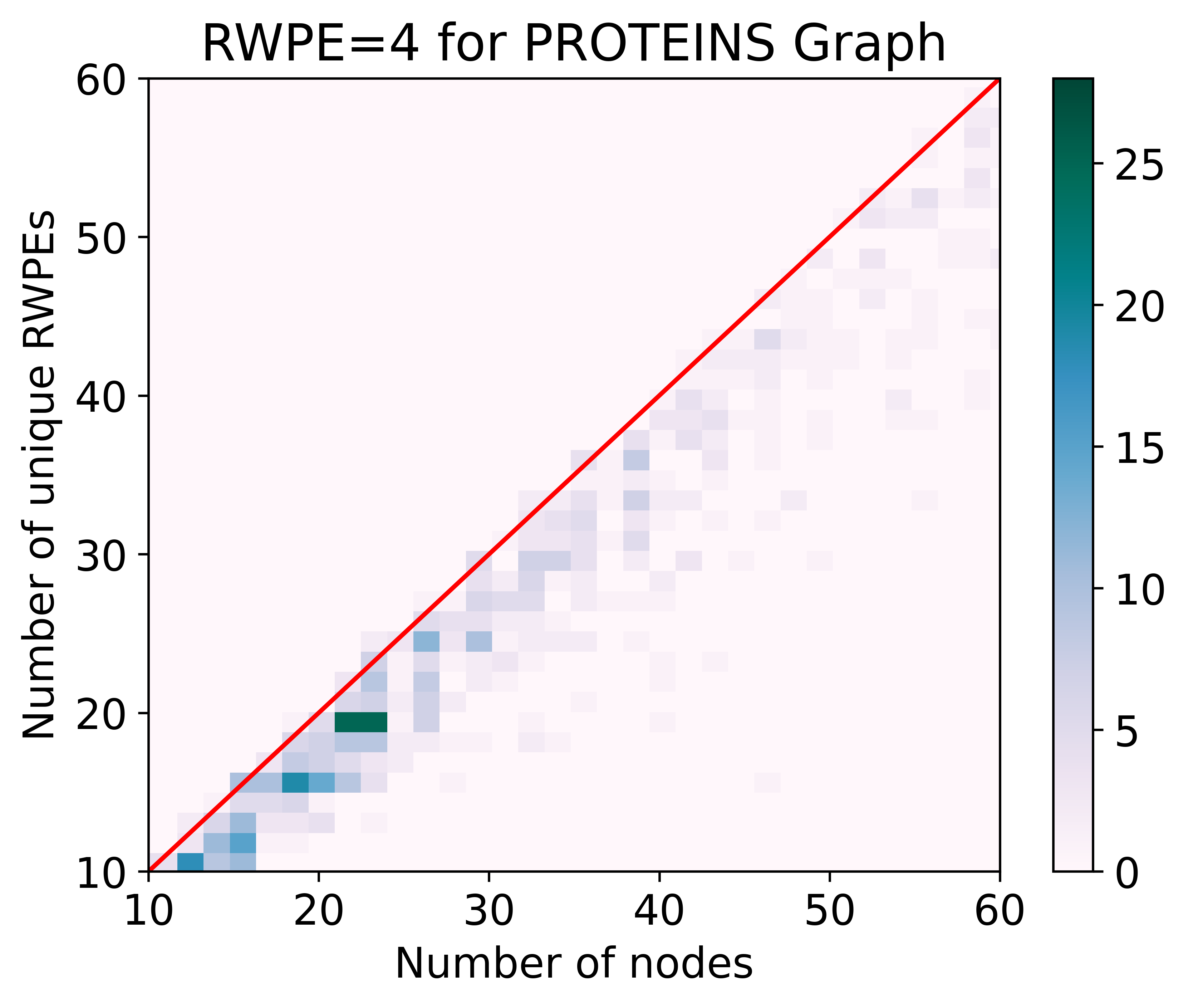}
    \label{fig:protein_rwpe4}
    \includegraphics[width=.32\columnwidth]{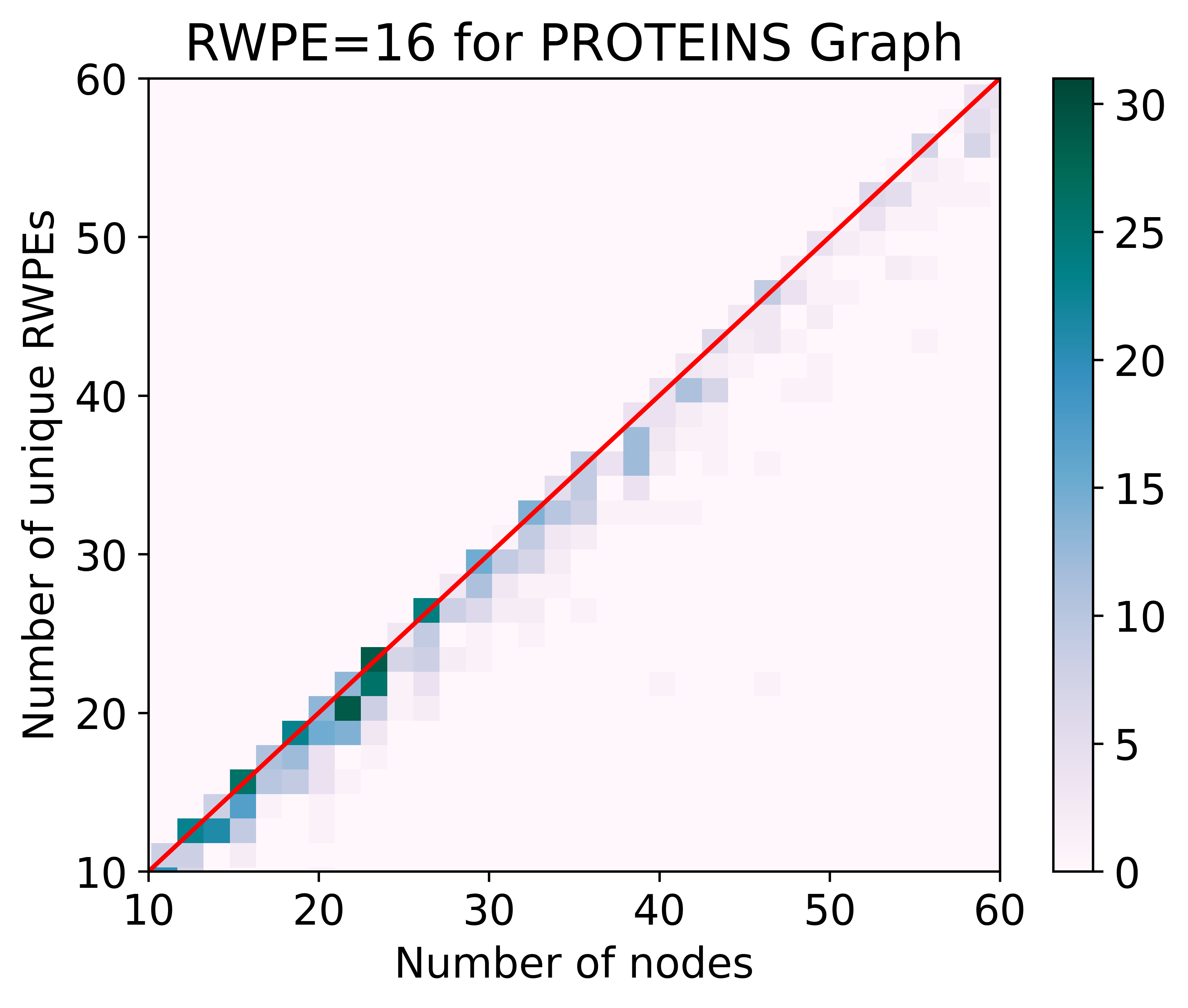}
    \label{fig:protein_rwpe16}
\caption{
\cd{The figure indicates how many nodes are distinguished out of the total number of nodes in the PROTEINS graphs when 1-WL or $RWPE$ is used. The x-axis is the total number of nodes in each graph, and the color on the plot is the number of graphs. Note that having the same number of nodes and unique nodes does not mean that the graphs are isomorphic, only that they can have the same number of (unique) nodes. }
}
\label{fig:unique_visualization}
\end{figure}
\cd{
Inspired by \cite{dwivedi2022graph}, we present a concrete analysis of the choice of $RWPE$ as a treatment for learning complex graphs compared to GNNs. In order to learn a complex structure like PROTEINS \cite{debnath1991structure}, the nodes in the graph must be distinguished by the correct unique positions, so that they are distinct and form a different graph when viewed as a whole. The Figure \ref{fig:unique_visualization} illustrates the number of uniquely distinct nodes in each graph of PROTEINS when distinguished by 1-WL and $RWPE$. A point on the plot indicates how many of the nodes are uniquely distinguished relative to the total nodes, which means the number of graphs at that point. The more graphs that are close to the red axis, the more uniquely distinguished nodes they have, and the more expressive they are. On the left side of Figure \ref{fig:unique_visualization}, we can see that the graphs distinguished by 1-WL are close to the red axis. Also, the $RWPE$ with $t=4$ in the middle is also close to the red axis, similar to 1-WL. This means that 1-WL and low-dimensional RWPE have similar expressive power in distinguishing graphs. When $RWPE$ is set to $t=16$, it is much closer to the red axis. To summarize, we can see that the choice of RWPE is more expressive compared to GNN, which is as expressive as 1-WL. The choice of a suitable dimension of $RWPE$ is more expressive compared to GNNs, which is as expressive as 1-WL. Thus, by utilizing $RWPE$, GPIO has effective expressivity for graph-related tasks.}

\section{Conclusion}

Multimodal learning graph structured dataset, as well as text, image, and speech, is necessary for general perception. In this paper, we propose GPIO(+) that handles the graph structured dataset as well as other datset such as image and text.

The strengths of our work is that we show that graph data, along with other modalities, can be effectively trained on a single architecture without the need for additional inductive bias.
This means that GPIO(+) learns topological information from positional encoding and output query design in the same way as other modalities, without changing the structure of the model to specialize in a particular modality.
\cf{These approaches demonstrates that GPIO is unaffected by the number of graph edges, achieving around 73\% memory savings over a typical GNN on a BA graph (edge density 0.03) with around 8000 nodes.}
For further research, there is potential to extend to the study of canonical positional embeddings on domain specific graphs and various graph-related multimodal learning with output query design.
\cf{However, our work does not scale well to domains where edge features are crucial, such as social graphs with multiple features like temporal dynamics and relationship types. It is necessary to propose a general architecture by adapting output query smoothing to incorporate edge features.}

\cf{
GPIO achieves higher accuracy on the PROTEIN dataset for graph classification (76.1\%) compared to DiffPool (74.3\%), and surpasses APPNP on PubMed (79.6\% vs. 79.1\%) for node classification. It also outperforms VGNAE on Cora (96.4\% vs. 95.7\%) in link prediction and improves upon RevGAT on ogbn-arxiv (76.14\% vs. 75.23\%) for multimodal text classification. Finally, GPIO+ achieves 76.1\% accuracy in multimodal few-shot classification, exceeding EGNN’s 73.1\%.
}

\cf{Overall, GPIO(+) is a general architecture for handling diverse modalities and tasks, including graph-related works, and can be applied to fields like recommendation systems and information retrieval as it accommodates graph, text, and image modalities.}



\section*{Declaration of Competing Interest}
There is no conflict of interest.

\section*{Data availability}
All of the code and data will be available upon request.

\section*{Acknowledgments}
This work was supported by the National Research Foundation of Korea(NRF) grant funded by the Korea government(MSIT) (No. 2022R1A4A3033874), and supported by the National Research Foundation of Korea(NRF) grant funded by the Korea government(MSIT). (No.2021R1F1A1060117).







\section{Appendix}

\subsection{Experimental Settings}
In this section, we describe the hyperparameter settings for each task and dataset. We set the same hyperparameter search spaces for each model in order to compare results fairly.

\subsubsection{Learning for Node Classification tasks}

For node classification tasks, we set the size of the output query array to $(M \times D_q)$. $M$ and $D_q$ are number of nodes and size of the learnable embedding vector respectively. The output array from the last cross attention layer is passed through the logits layer to make the vector size of each node equal to the class size. The size of the output array is, then, ($M \times E$). $E$ should be the same size with number of classes. We evaluate the cross-entropy loss for node classification tasks.
\begin{eqnarray}\label{eq:node_loss}
    \mathcal{L}_{C.E} = -\frac{1}{M}\sum_{m=1}^{M}log(\frac{w_{my}}{\sum_{e=1}^{E}w_{me}})
\end{eqnarray}
$w_{my}$ and $w_{me}$ denote the value of element of $m$ th node's row vector corresponding to the label index in output array and each element value of $m$ th node's row vector respectively.

\subsubsection{Learning for Link Prediction tasks}

We use the GPIO as an encoder to embed the raw node features to the latent embeddings and, we adopt an inner product to predict edges. Specifically, the output array generated by our model becomes a latent embedding matrix. In the same manner, as GAE \cite{kipf2016variational}, we conduct the inner product between each pair of node features. 
\begin{eqnarray}\label{eq:link_pred_1}
    P(\mathbf{A}|\mathbf{W}) = \prod_{i=1}^{N}\prod_{j=1}^{N}P(A_{ij}|w_i, w_j),\\
    \centering\text{with} \, P(A_{ij}=1|w_i, w_j) = \sigma(w_i^T\cdot w_j) \nonumber
\end{eqnarray}
Matrix $\mathbf{A}$ denotes the adjacency matrix, and $\mathbf{W}$ is the output array. $\sigma$ denotes a sigmoid function. We use reconstruction loss for training the model for link prediction.
\begin{eqnarray}\label{eq:link_loss}
\mathcal{L}_{recon} = -\log(\frac{1}{|N^{pos}_{tr}|}\sum_{(i,j)\in N^{pos}_{tr}}\sigma(w_i^T\cdot w_j)) \nonumber \\
\; -\log(1-\frac{1}{|N^{neg}_{tr}|}\sum_{(k,t)\in N^{neg}_{tr}}\sigma(w_k^T\cdot w_t))
\end{eqnarray}
$N_{tr}^{pos}$ denotes a set of positive train edges. $N_{tr}^{neg}$ denotes a set of random sampled negative train edges. A $\sigma$ function maps the inner product of embeddings of node pairs to a proper value between 0 and 1. The reconstruction loss Eq. \ref{eq:link_loss} is binary cross-entropy loss, which assigns positive edges a label of one and negative edges a label of zero.

\subsubsection{Learning for Graph Classification tasks}
Lastly, for graph classification tasks, the output query array is 1-dimensional array. After through the logits layer, It still be a 1-dimensional array but the size of array should match the number of classes. The characteristic of this procedure is that we can conduct graph-level tasks without the read-out layer (e.g., global pooling layer). We use the cross-entropy loss generally used in graph classifications.
\begin{eqnarray}\label{eq:link_pred_2}
    \mathcal{L}_{C.E} = -\frac{1}{H}\sum_{h=1}^{H}log(\frac{w_{hy}}{\sum_{e=1}^{E}w_{he}})
\end{eqnarray}
H denotes the number of graphs. $w_{hy}$ and $w_{he}$ denote the value of element of $h$th graph's row vector corresponding to the label index in the output array and each element value of $h$th graph's row vector respectively.

\subsubsection{Hyperparameter Setting for Node Classification}
Table \ref{tab:hyp_node} describes the hyperparameter search spaces for the node classification task.

\begin{table}[h!]
\centering
\resizebox{0.65\textwidth}{!}{%
\begin{tabular}[t]{l|ccc}
\multicolumn{1}{c|}{\multirow{2}{*}{Hyperparameter}} & \multicolumn{3}{c}{Candidate values} \\
\multicolumn{1}{c|}{} & Cora & CiteSeer & PubMed \\ \hline
learning rate       & \{5e-4, 1e-3\} & \{1e-3\} & \{5e-3, 1e-2\} \\
weight decay        & \{1e-2\} & \{5e-2\} & \{5e-2, 1e-1\} \\
latent length       & \{16, 32, 64\} & \{4\} & \{4, 8\} \\
latent dimension    & \{32, 64\} & \{64, 128\} & \{32, 64, 128, 256\} \\
\# of MHCA heads  & \{32, 64\} & \{32, 64\} & \{1\} \\
\# of MHSA heads  & \{8, 16\} & \{8, 16 32\} & \{1, 8, 16\} \\
MHCA head dimension  & \{64, 128\} & \{32, 64, 128\} & \{64, 128\} \\
MHSA head dimension  & \{64, 128\} & \{32, 64, 128\} & \{64, 128\} \\
depth  & \{1\} & \{1\} & \{1\} \\ \hline
\end{tabular}
}
\caption{Hyperparameters for the node classification task. MHCA and MHSA denote the multi-head cross attention and multi-head self attention, respectively.}
\label{tab:hyp_node}
\end{table}

\subsubsection{Hyperparameter Setting for Link Prediction}
Table \ref{tab:hyp_link} describes the hyperparameter search spaces for the link prediction task.

\begin{table}[h]
\centering
\resizebox{0.6\textwidth}{!}{%
\begin{tabular}[t]{l|ccc}
\multicolumn{1}{c|}{\multirow{2}{*}{Hyperparameter}} & \multicolumn{3}{c}{Candidate values} \\
\multicolumn{1}{c|}{} & Cora & CiteSeer & PubMed \\ \hline
learning rate       & \{5e-5\} & \{1e-5\} & \{5e-4\} \\
weight decay        & \{5e-4, 1e-3\} & \{5e-3, 1e-2\} & \{5e-4, 1e-3\} \\
latent length       & \{16, 32\} & \{32, 64\} & \{32, 64\} \\
latent dimension    & \{32, 64\} & \{256,512\} & \{32, 64\} \\
\# of MHCA heads  & \{64, 128\} & \{128, 256\} & \{16, 32\} \\
\# of MHSA heads  & \{8, 16\} & \{4, 8\} & \{4, 8\} \\
MHCA head dimension  & \{128, 256\} & \{128, 256\} & \{128, 256\} \\
MHSA head dimension  & \{128, 256\} & \{128, 256\} & \{128, 256\} \\
depth  & \{1\} & \{1\} & \{1\} \\ \hline
\end{tabular}
    }
\caption{Hyperparameters for the link prediction task. MHCA and MHSA denote the multi-head cross attention and multi-head self attention, respectively.}
\label{tab:hyp_link}
\end{table}


\subsubsection{Hyperparameter Setting for Graph Classification}
Table \ref{tab:hyp_graph} describes the hyperparameter search spaces for the graph classification task.

\begin{table}[hbt!]
\centering
\resizebox{1.0\textwidth}{!}{%
\begin{tabular}[t]{l|ccccc}
\multicolumn{1}{c|}{\multirow{2}{*}{Hyperparameter}} & \multicolumn{5}{c}{Candidate values} \\
\multicolumn{1}{c|}{} & MUTAG & PROTEINS & IMDB-BINARY & REDDIT-BINARY & COLLAB     \\ \hline
learning rate       & \{5e-3, 1e-3, 5e-4\} & \{5e-3, 1e-3, 5e-4\} & \{5e-3, 1e-3\} & \{5e-3\}& \{5e-3, 1e-3\}\\
weight decay        & \{5e-4, 1e-4\} & \{5e-4, 1e-4\} & \{5e-4, 1e-4\} & \{5e-4, 1e-4\} & \{5e-4, 1e-4\} \\
latent length       & \{8, 16, 32\} & \{16, 32, 64\} & \{32, 64, 128\} & \{32, 64\} & \{64\} \\
latent dimension    & \{32, 64, 128\} & \{32, 64, 128\} & \{32, 64, 128, 256\} & \{32, 64\} & \{256\} \\
\# of MHCA heads  & \{1\} & \{1\} & \{1\} & \{8, 16\} & \{8, 16\} \\
\# of MHSA heads  & \{1\} & \{1\} & \{1\} & \{4\} & \{4\} \\
MHCA head dimension  & \{4\} & \{4\} & \{4\} & \{4\} & \{4\} \\
MHSA head dimension  & \{4\} & \{4\} & \{4\} & \{4\} & \{4\} \\
depth  & \{1\} & \{1\} & \{1\} & \{1\} & \{1\} \\ \hline
\end{tabular}
}
\caption{Hyperparameters for the graph classification task. MHCA and MHSA denote the multi-head cross attention and multi-head self attention, respectively.}
\label{tab:hyp_graph}
\end{table}

\subsubsection{Hyperparameter setting for Multimodal Few-shot Classification}
Table \ref{tab:hyp-fs} describes the hyperparameter search spaces for the multimodal few-shot classification task.

\begin{table}[hbt!]
\centering
\resizebox{0.35\textwidth}{!}{%
\begin{tabular}{l|c}
\multicolumn{1}{c|}{Hyperparameter} & Candidate values  \\ \hline
learning rate                       & \{5e-6, 1e-5\}    \\
weight decay                        & \{0, 1e-6, 1e-5\} \\
latent length                       & \{32\}            \\
latent dimension                    & \{32, 64\}        \\
\# of MHCA heads                     & \{16, 64\}        \\
\# of MHSA heads                     & \{4\}             \\
MHCA head dimension                 & \{256, 512\}      \\
MHSA head dimension                 & \{256, 512\}      \\
depth                               & \{1, 2\}          \\ \hline
\end{tabular}%
}
\caption{Hyperparameters for the multimodal few-shot classification task. MHCA and MHSA denote the multi-head cross attention and multi-head self attention, respectively.}
\label{tab:hyp-fs}
\end{table}

\subsubsection{\cf{Hyperparameter setting for Multimodal Text Classification}}
\cf{Table \ref{tab:hyp-tc} describes the hyperparameter search spaces for the multimodal text classification task.}

\begin{table}[hbt!]
\centering
\resizebox{0.45\textwidth}{!}{%
\begin{tabular}{l|cc}
\multicolumn{1}{c|}{\multirow{2}{*}{Hyperparameter}} & \multicolumn{2}{c}{Candidate values} \\
\multicolumn{1}{c|}{} & ogbn-products & ogbn-arxiv  \\ \hline
learning rate       & \{1e-3\} & \{5e-4\}  \\
weight decay        & \{5e-4\} & \{5e-4\} \\
latent length       & \{16, 32, 64\} & \{16, 32, 64\}  \\
latent dimension    & \{128, 256\} & \{128, 256\} \\
\# of MHCA heads  & \{8, 16\} & \{8, 16\} \\
\# of MHSA heads  & \{8, 16, 32\} & \{8, 16 32\} \\
MHCA head dimension  & \{64, 128\} & \{64, 128\}  \\
MHSA head dimension  & \{64, 128\} & \{64, 128\}  \\
depth  & \{1\} & \{1\}  \\ \hline
\end{tabular}%
}
\caption{\cf{Hyperparameters for the multimodal text classification task. MHCA and MHSA denote the multi-head cross attention and multi-head self attention, respectively.}}
\label{tab:hyp-tc}
\end{table}

\subsubsection{Additional Experimental Settings}
\cf{For all classification tasks except link prediction, we used the accuracy metric to measure whether the node, graph, text, or image is correctly classified.
For link prediction, a binary classification task, we adopt average precision (AP) and area under the ROC curve (AUC) to effectively handle class imbalance, as non-existent edges are more frequent than existing ones.}

We validate our models on RTX A6000, RTX3090, and A100 GPU. We adopt random seed value 2025 for all experiments. We use PyTorch 1.11.0 and PyTorch-Geometric 2.0.1 with CUDA 11.2.

\subsection{Additional Experimental Results}
In this section, we provide additional quantitative results.

\subsubsection{Link Prediction}

\begin{figure}[h]
    \centering
    \includegraphics[width=.27\columnwidth]{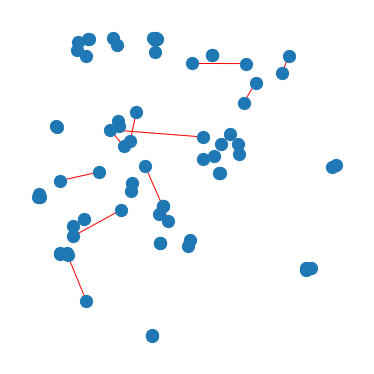}
    \label{fig:pos_io}
    \includegraphics[width=.27\columnwidth]{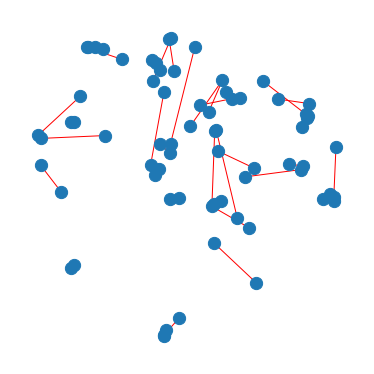}
    \label{fig:pos_gnae}
    \includegraphics[width=.27\columnwidth]{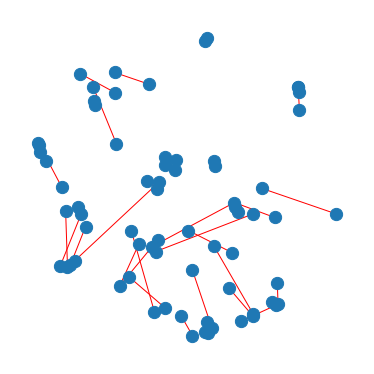}
    \label{fig:pos_vgnae}
    \newline
    \centering
    \includegraphics[width=.27\columnwidth]{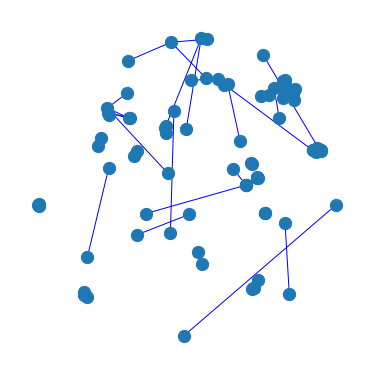}
    \label{fig:neg_io}
    \includegraphics[width=.27\columnwidth]{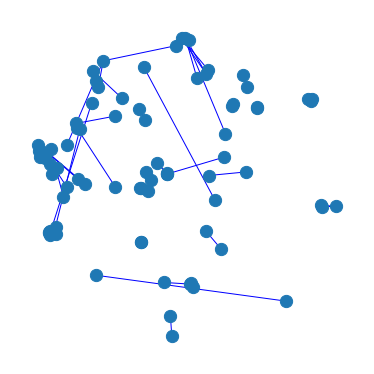}
    \label{fig:neg_gnae}
    \includegraphics[width=.27\columnwidth]{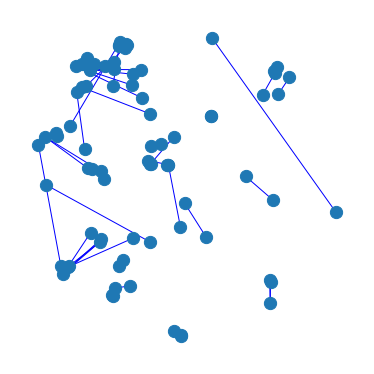}
    \label{fig:neg_vgnae}
\caption{
The first row denotes the t-SNE of positive edges (red color) leanred by GPIO (left), GNAE (middle), VGNAE (right). The second row denotes the t-SNE of negative edges (blue color) learned by GPIO (left), GNAE (middle), VGNAE (right). Compared to other models, our model embeds the node pairs of the positive edge set closer together, indicating that the node representation is better in the edge level task. Contrary, in the negative edge case, the model is more expressive when the distance between two nodes increases.
}
\label{fig:edge_tsne}
\end{figure}

\begin{table}[t!]
\centering
\resizebox{1\textwidth}{!}{\begin{tabular}{llcccccccc}
\hline
\multicolumn{1}{c|}{\multirow{2}{*}{Model}}    &\multicolumn{1}{c|}{\multirow{2}{*}{$RWPE$}}&\multicolumn{1}{c|}{Output query}                         & \multicolumn{2}{c}{Cora}                                  & \multicolumn{2}{c}{CiteSeer}                             & \multicolumn{2}{c}{PubMed}                                \\
\multicolumn{1}{l|}{}                          &\multicolumn{1}{c|}{}                      &\multicolumn{1}{c|}{smoothing}                           & AUC                         & AP                          & AUC                         & AP                         & AUC                        & AP                           \\ \hline
\multicolumn{1}{l|}{Spectral Clustering \cite{tang2011leveraging}*} &\multicolumn{1}{c|}{} &\multicolumn{1}{l|}{}                                    & 84.6$\pm{0.01}$             & 88.5$\pm{0.0}$              & 80.5$\pm{0.01}$             & 85.0$\pm{0.01}$            & 84.2$\pm{0.02}$            & 87.8$\pm{0.01}$              \\
\multicolumn{1}{l|}{DeepWalk \cite{perozzi2014deepwalk}*}           &\multicolumn{1}{c|}{} &\multicolumn{1}{l|}{}                                    & 83.1$\pm{0.01}$             & 85.0$\pm{0.0}$              & 80.5$\pm{0.02}$             & 83.6$\pm{0.01}$            & 84.4$\pm{0.0}$             & 84.1$\pm{0.0}$               \\
\multicolumn{1}{l|}{DGI \cite{velickovic2019deep}*}                 &\multicolumn{1}{c|}{} &\multicolumn{1}{l|}{}                                    & 89.8$\pm{0.8}$              & 89.7$\pm{1.0}$              & 95.5$\pm{1.0}$              & 95.7$\pm{1.0}$             & 91.2$\pm{0.6}$             & 92.2$\pm{0.5}$               \\
\multicolumn{1}{l|}{ARGVA \cite{pan2019learning}*}                  &\multicolumn{1}{c|}{} &\multicolumn{1}{l|}{}                                    & 92.4$\pm{0.004}$            & 93.2$\pm{0.003}$            & 92.4$\pm{0.003}$            & 93.0$\pm{0.003}$           & 96.8$\pm{0.001}$           & 97.1$\pm{0.001}$             \\
\multicolumn{1}{l|}{GIC \cite{DBLP:conf/pakdd/MavromatisK21}}                 &\multicolumn{1}{c|}{} &\multicolumn{1}{l|}{}                                    & 90.0$\pm{1.0}$              & 89.9$\pm{1.3}$              & 95.8$\pm{0.7}$              & 95.8$\pm{0.9}$             & 90.9$\pm{1.0}$             & 91.6$\pm{0.9}$               \\
\multicolumn{1}{l|}{VGAE \cite{kipf2016variational}}                &\multicolumn{1}{c|}{} &\multicolumn{1}{l|}{}                                    & 95.2$\pm{0.5}$              & 94.7$\pm{0.6}$              & 92.0$\pm{1.7}$              & 91.6$\pm{1.7}$             & 95.6$\pm{0.7}$             & 95.3$\pm{0.6}$               \\
\multicolumn{1}{l|}{GNAE \cite{ahn2021variational}}                 &\multicolumn{1}{c|}{} &\multicolumn{1}{l|}{}                                   & 95.6$\pm{0.7}$           & 96.0$\pm{0.8}$  & \textbf{97.2}$\pm{0.5}$  & \textbf{97.3}$\pm{0.4}$  & \textbf{97.7}$\pm{0.2}$  & \underline{97.6}$\pm{0.2}$        \\
\multicolumn{1}{l|}{VGNAE \cite{ahn2021variational}}                &\multicolumn{1}{c|}{} &\multicolumn{1}{l|}{}                                     & 95.8$\pm{0.6}$  & 95.7$\pm{0.8}$           & 96.8$\pm{0.6}$           & 96.7$\pm{0.6}$           & 97.3$\pm{0.1}$           & 97.2$\pm{0.2}$             \\ \hline
\multicolumn{1}{l|}{\multirow{5}{*}{GPIO}}            &\multicolumn{1}{c|}{-}&\multicolumn{1}{c|}{$L=0$}   & 94.4$\pm{0.3}$	           &95.4	$\pm{0.3}$	         &91.5$\pm{0.3}$	           &91.1$\pm{0.7}$	            &95.9$\pm{0.1}$	         &95.9$\pm{0.09}$               \\
\multicolumn{1}{l|}{}                                               &\multicolumn{1}{c|}{\checkmark} &\multicolumn{1}{c|}{$L=0$}   &94.3$\pm{0.4}$	           &95.2	$\pm{0.3}$	         &89.3$\pm{0.7}$	           &91.0$\pm{0.5}$	            &95.8$\pm{0.1}$	             &95.7$\pm{0.1}$                \\
\multicolumn{1}{l|}{}                                               &\multicolumn{1}{c|}{-} &\multicolumn{1}{c|}{$L=1$}          & 95.6$\pm{0.2}$              & \underline{96.3}$\pm{0.1}$             & 96.2$\pm{0.3}$              & 96.7$\pm{0.2}$              & \underline{97.6}$\pm{0.08}$                & \textbf{97.8}$\pm{0.07}$                                                      \\ 
\multicolumn{1}{l|}{}                                               &\multicolumn{1}{c|}{\checkmark} &\multicolumn{1}{c|}{$L=1$}         & \textbf{95.9}$\pm{0.3}$	           &\textbf{96.4}	$\pm{0.2}$	         &95.6$\pm{0.5}$	           &96.1$\pm{0.3}$	            &97.6$\pm{0.06}$	         &97.7$\pm{0.08}$               \\
\multicolumn{1}{l|}{}                                               &\multicolumn{1}{c|}{-} &\multicolumn{1}{c|}{$L=2$}          &\underline{95.8}$\pm{0.2}$	           &96.3	$\pm{0.1}$	         &\underline{96.8}$\pm{0.3}$              &\underline{97.2}$\pm{0.2}$	            &97.0$\pm{0.09}$	         &97.2$\pm{0.08}$               \\ \hline
\end{tabular}}
\caption{The performance of link prediction task. The GPIO shows competitive performance compared with the state-of-the-art model, \cite{ahn2021variational}. Also, the GPIO shows the superior results with the well-known link prediction model VGAE, which adopts the graph convolutional neural networks. * denotes the reported performance in \cite{DBLP:conf/pakdd/MavromatisK21}. We use 4 dimensional $RWPE$ in link prediction. $L=0, 1, 2$ denotes number of smoothing.}
\label{tab:line_prediction_supple}
\end{table}

\cd{
We analyze the impact of the smoothing times $L$ and the effect of $RWPE$ on link prediction. Table \ref{tab:line_prediction_supple} compares performance to the baseline, showing how our model performs as $L$ varies from 0 to 2 and based on the presence or absence of $RWPE$. Feature smoothing that reflects relational information performs better than without it. In the Link prediction task, we can see that the presence of $RWPE$ does not have a significant impact on performance, as enough relational information is captured by feature smoothing. }

\subsubsection{Node Classification}

\begin{table}[h!]
\centering
\resizebox{1\textwidth}{!}{%
\footnotesize
\begin{tabular}{lllccccccc}
\hline
\multicolumn{1}{c|}{\multirow{2}{*}{Model}}                     &\multicolumn{1}{c|}{\multirow{2}{*}{$RWPE$}} &\multicolumn{1}{c|}{Output query}         & \multicolumn{2}{c}{Cora}                                 & \multicolumn{2}{c}{CiteSeer}  & \multicolumn{2}{c}{PubMed}    \\
\multicolumn{1}{l|}{}                                          &\multicolumn{1}{l|}{}                     &\multicolumn{1}{c|}{smoothing}            & Fixed                       & Random                     & Fixed                      & Random        & Fixed         & Random        \\ \hline
\multicolumn{1}{l|}{GCN \cite{kipf2017semisupervised}}                   &\multicolumn{1}{l|}{}                     &\multicolumn{1}{l|}{}                    & 81.4$\pm{0.7}$              & 78.7$\pm{1.7}$             & 71.1$\pm{0.7}$             & 68.1$\pm{1.7}$              & 78.9$\pm{0.6}$          & 77.3$\pm{2.4}$          \\
\multicolumn{1}{l|}{GAT \cite{velivckovic2017graph}}           &\multicolumn{1}{l|}{}                     &\multicolumn{1}{l|}{}                    & 83.0$\pm{0.5}$          & 80.9$\pm{1.5}$          & 70.9$\pm{0.5}$          & \underline{68.8}$\pm{1.7}$          & 78.9$\pm{0.4}$         & 77.6$\pm{2.4}$          \\
\multicolumn{1}{l|}{Cheb \cite{defferrard2016convolutional}}   &\multicolumn{1}{l|}{}                     &\multicolumn{1}{l|}{}                    & 80.5$\pm{1.0}$              & 77.0$\pm{2.7}$             & 69.8$\pm{1.2}$             & 67.2$\pm{2.1}$              & 78.2$\pm{0.6}$          & 75.4$\pm{2.5}$          \\
\multicolumn{1}{l|}{SGC \cite{wu2019simplifying}}              &\multicolumn{1}{l|}{}                     &\multicolumn{1}{l|}{}                    & 81.7$\pm{0.05}$             & 80.1$\pm{1.9}$             & \underline{71.3}$\pm{0.2}$ & 68.5$\pm{2.0}$              & 78.9$\pm{0.0}$          & 76.6$\pm{2.4}$          \\
\multicolumn{1}{l|}{ARMA \cite{bianchi2021graph}}              &\multicolumn{1}{l|}{}                     &\multicolumn{1}{l|}{}                    & 82.2$\pm{0.9}$              & 79.8$\pm{1.7}$             & 71.0$\pm{0.6}$             & 67.9$\pm{1.9}$              & 78.8$\pm{0.3}$          & 77.6$\pm{2.2}$          \\
\multicolumn{1}{l|}{APPNP \cite{gasteiger2018combining}}          &\multicolumn{1}{l|}{}                     &\multicolumn{1}{l|}{}                    & \underline{83.3}$\pm{0.5}$ & \textbf{82.1$\pm{1.5}$} & \textbf{71.7$\pm{0.5}$} & \textbf{69.8$\pm{1.8}$} & \textbf{80.1$\pm{0.2}$} & \underline{79.1}$\pm{2.3}$          \\ \hline
\multicolumn{1}{l|}{\multirow{6}{*}{\begin{tabular}[c]{@{}l@{}}GPIO\end{tabular}}}       &\multicolumn{1}{c|}{-}                    &\multicolumn{1}{l|}{$L=0$}        &59.3	$\pm{1.2}$		       &57.9	$\pm{2.6}$		    &57.4	$\pm{1.3}$		     &55.0	$\pm{2.3}$	      	&72.3	$\pm{1.7}$		     &70.0	$\pm{3.0}$ \\
\multicolumn{1}{l|}{}             	                           &\multicolumn{1}{c|}{\checkmark}                    &\multicolumn{1}{l|}{$L=0$} &58.6	$\pm{1.9}$		       &57.4	$\pm{2.3}$		    &57.4	$\pm{1.5}$		     &55.5	$\pm{2.5}$		     &73.0	$\pm{1.1}$		     &70.0	$\pm{2.6}$ \\
\multicolumn{1}{l|}{}                 	                       &\multicolumn{1}{c|}{-}                    &\multicolumn{1}{l|}{$L=2$}            &81.8	$\pm{0.7}$		       &79.9	$\pm{1.8}$		    &69.3	$\pm{1.2}$		     &67.6	$\pm{1.8}$		     &78.6	$\pm{0.5}$		     &77.9	$\pm{2.4}$ \\
\multicolumn{1}{l|}{}                 	                       &\multicolumn{1}{c|}{-}                    &\multicolumn{1}{l|}{$L=3$}             &82.9	$\pm{0.7}$		       &80.7	$\pm{1.6}$		    &68.9	$\pm{1.1}$		     &67.7	$\pm{1.9}$		     &78.2	$\pm{0.4}$	     	&78.0	$\pm{2.4}$ \\
\multicolumn{1}{l|}{}                 	                       &\multicolumn{1}{c|}{-}                    &\multicolumn{1}{l|}{$L=10, \alpha=0.1$}          &\textbf{83.9}	$\pm{0.6}$		       &\underline{81.6}	$\pm{1.8}$		    &70.1	$\pm{1.0}$		     &68.1	$\pm{1.7}$	     	&79.8	$\pm{0.5}$		     &\textbf{79.6$\pm{2.2}$} \\
\multicolumn{1}{l|}{}                 	                       &\multicolumn{1}{c|}{\checkmark}                    &\multicolumn{1}{l|}{$L=10, \alpha=0.1$}   &83.7	$\pm{0.8}$		       &81.5	$\pm{1.5}$		    &69.2	$\pm{1.0}$		     &67.9	$\pm{1.7}$	       	&\underline{79.9}$\pm{0.4}$	  &79.0	$\pm{2.5}$  \\\hline
\end{tabular}
}
\caption{Accuracy for the node classification. The GPIO shows the competitive results on the PubMed. Fixed and Random denotes the dataset split methods from \cite{Fey:2019wv}. We use 4 or 8 dimensional $RWPE$ in node classification. $L=0, 2, 3$ denotes number of smoothing, $L=10, \alpha=0.1$ denotes hyperparameters of APPNP.}
\label{tab:noderesults_supple}
\end{table}

\cd{
We analyzed the impact of the output query array smoothing type and its hyperparameters on node classification and the effect on $RWPE$. We find that the output query array smoothed by APPNP shows comparable performance with the GNNs. Since $\widehat{X}_{APPNP}$ has less oversmoothing, we can set the $L$ value to be larger than $\widehat{X}$, and it has the effect of propagating the score $\widehat{S}_{\mathit{APPNP}}$ to more distant nodes and is effective in node classification. On the other hand, as with link prediction, $RWPE$ does not have a significant impact, as the relational information is well reflected by $\widehat{X}_{APPNP}$.}

\begin{figure}[!h]
\centering
\includegraphics[width=.5\columnwidth]{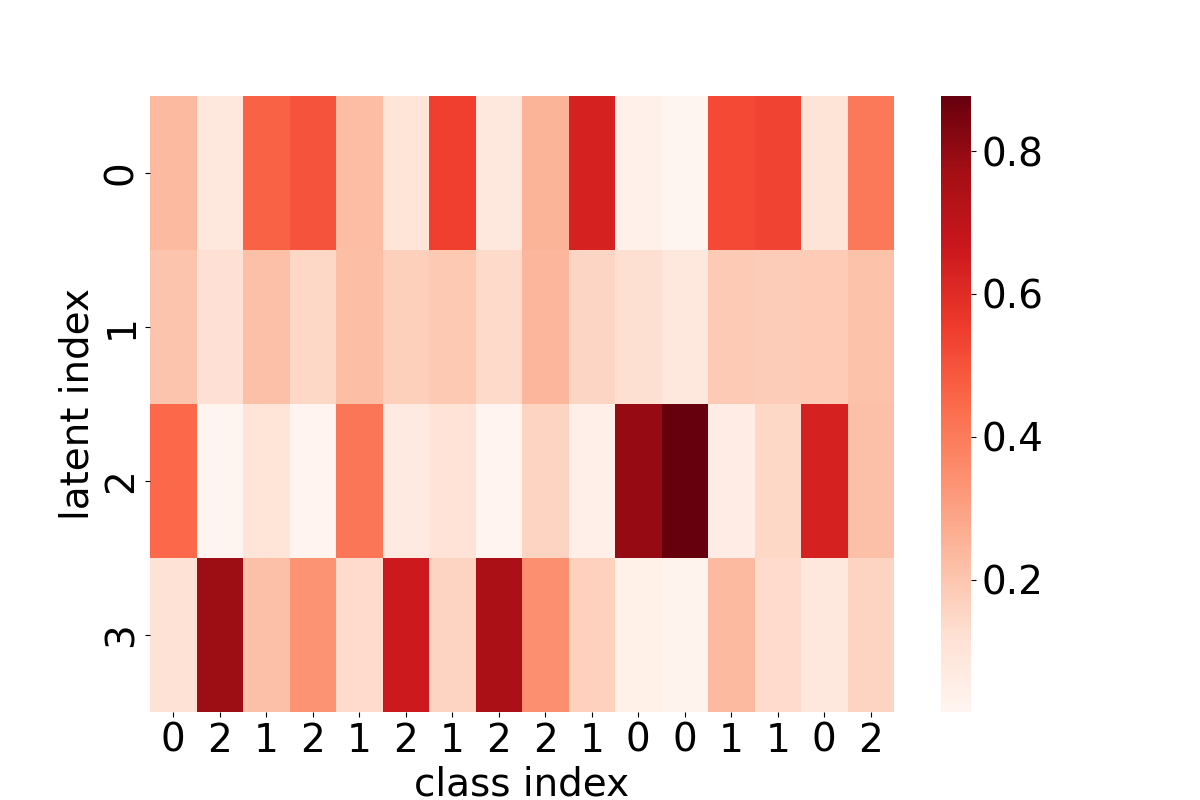}
\caption{GPIO attention heatmap for each latent index on PubMed. The x-axis and y-axis indicate the node class and latent index, respectively. Each latent of the GPIO captures the different nodes with diversity. Besides, each latent prone to focus on the specific class.}
\label{fig:node_attention_heatmap}
\end{figure}

To analyze the attention properties of the GPIO, we visualize the attention weights of cross-attention and the attention between the latent arrays and the input arrays as shown in Figure \ref{fig:node_attention_heatmap}. Each latent index in the GPIO captures the different nodes with high diversity. Additionally, each latent focuses on the specific class group of nodes relatively.

\subsubsection{Graph Classification}

\begin{table}[h!]
\centering
\resizebox{0.9\textwidth}{!}{%
\begin{tabular}{lccccc}
\hline
\multicolumn{1}{c}{Model} & MUTAG         & PROTEINS      & \begin{tabular}[c]{@{}c@{}}IMDB-\\ BINARY\end{tabular} & \begin{tabular}[c]{@{}c@{}}REDDIT-\\ BINARY\end{tabular} & COLLAB        \\ \hline
GCNWithJK                 & 72.9$\pm{12.0}$          & 72.6$\pm{3.6}$          & 73.2$\pm{5.0}$                                                   & 89.4$\pm{2.9}$                                                     & \textbf{81.5}$\pm{2.1}$ \\
SAGPool                   & 74.0$\pm{8.7}$          & 72.3$\pm{2.8}$          & 72.3$\pm{4.7}$                                                   & 89.0$\pm{2.1}$                                                     & 78.9$\pm{1.0}$          \\
DiffPool                  & \underline{84.6}$\pm{8.7}$          & \underline{74.3}$\pm{6.4}$          & \textbf{74.8}$\pm{4.8}$                                          & \textbf{92.7}$\pm{2.0}$                                            & 79.4$\pm{1.9}$          \\
EdgePool                  & 72.3$\pm{13.4}$          & 72.6$\pm{3.4}$          & 73.3$\pm{5.3}$                                                   & 89.2$\pm{3.8}$                                                     & 79.3$\pm{1.2}$          \\
GCN                       & 70.7$\pm{11.0}$          & 72.2$\pm{2.4}$          & 74.2$\pm{4.4}$                                                   & 89.1$\pm{2.0}$                                                     & \underline{81.0}$\pm{1.4}$          \\
GraphSAGE                 & 75.1$\pm{11.4}$          & 74.1$\pm{2.3}$          & 73.2$\pm{4.4}$                                                   & 90.7$\pm{2.3}$                                                     & 79.8$\pm{1.1}$          \\
GIN0                      & 81.9$\pm{8.0}$          & 73.1$\pm{3.8}$          & \underline{73.7}$\pm{4.1}$                                                   & 90.9$\pm{2.1}$                                                     & 80.5$\pm{1.9}$          \\
GIN                       & 82.9$\pm{11.3}$          & 72.0$\pm{3.2}$            & 73.5$\pm{4.9}$                                                   & 90.5$\pm{2.5}$                                                     & 80.7$\pm{2.0}$          \\
GlobalAttentionNet        & 77.7$\pm{12.1}$          & 72.6$\pm{2.6}$          & 72.9$\pm{3.7}$                                                   & 88.6$\pm{3.2}$                                                     & 79.1$\pm{0.7}$          \\
Set2SetNet                & 74.5$\pm{11.9}$          & 74.1$\pm{3.8}$          & 72.7$\pm{5.0}$                                                   & 90.3$\pm{2.4}$                                                     & 79.4$\pm{1.7}$          \\
SortPool                  & 83.0$\pm{9.0}$          & 73.9$\pm{4.5}$          & 71.8$\pm{3.0}$                                                   & 84.3$\pm{5.0}$                                                     & 77.8$\pm{1.6}$          \\
ASAP                      & 78.7$\pm{11.8}$          & 74.0$\pm{3.0}$            & 72.2$\pm{4.3}$                                                   & OOM                                                      & 79.4$\pm{1.7}$          \\ \hline
Graph Percevier IO        &                         &                          &                                                                  &                                                                        &                      \\
\multicolumn{1}{l}{+ None PE}	      &69.6$\pm{6.6}$	         &71.1$\pm{3.7}$	       &72.0$\pm{5.2}$	                                                   &86.4$\pm{2.9}$	                                                    &77.7$\pm{2.4}$ \\
\multicolumn{1}{l}{+ Fourier PE}	  &83.6$\pm{9.5}$	           &72.5$\pm{4.3}$	       &71.7$\pm{4.3}$	                                                   &76.1$\pm{3.5}$	                                                    &75.5$\pm{2.0}$ \\ 
\multicolumn{1}{l}{+ $RWPE$}        & \textbf{86.1}$\pm{6.9}$ & \textbf{76.1}$\pm{3.0}$ & 72.9$\pm{4.4}$                                                   & \underline{91.6}$\pm{2.3}$                                                     & 79.1$\pm{2.5}$          \\ \hline
\end{tabular}
}
\caption{Accuracy of the graph classification tasks. The GPIO shows the superior performance on MUTAG and PROTEINS dataset that has node features. IMDB, REDDIT, and COLLAB have no node features, and an understanding of the topological information of the graph is required. The GPIO shows the competitive results even though there are no node features. We use 64 dimensional $RWPE$ in graph classification. For None PE in the REDDIT dataset, we remove first layer normalization.}
\label{tab:graph_classification_supple}
\end{table}
\cd{
Table \ref{tab:graph_classification_supple} denotes how the presence and type of positional encoding affects graph classification performance. Unlike node classification, GPIO uses an output query array of $1 \times D_q(=E) $ because the model classifies the entire graph rather than each node in graph classification. This means that unlike other tasks, it reflects topological information via $RWPE$ without output query array smoothing. 
The results show that on relatively small and simple MUTAG and PROTEINS datasets, Fourier positional encoding that incorporates only sequential information outperforms no positional encoding. However, in large, complex graphs in other datasets, sequential information caused overfitting. This is because a graph with a same structure seen at training time may have been encoded as a completely different graph at test time, depending on how the Fourier PE were started. In contrast, $RWPE$ outperformed None and Fourier PE on all datasets. This is because $RWPE$ effectively reflects the graph structure that the GPIO saw at training time to the same or similar graph structure at test time. This demonstrates  how effective $RWPE$ is at providing canonical positional information in graph classification. 
}

\subsection{Graph Related Tasks}

\begin{table}[h!]
\centering

\begin{tabular}{c|cccc}
Tasks                  & input array    & Output query array & Logits layer \\ \hline
Node classification          & $ M\times (C+t) $            & $M \times D_q(=C)$         & O           \\
Graph classification          & $ M\times (C+t) $     & $1 \times D_q(=E) $         & O           \\
Link prediction            & $ M\times (C+t) $            & $ M \times D_q(=C) $         & X           \\ 
\end{tabular}
\caption{Configuration of input and output query array used in the actual training process. We adopt a smoothing based output query array for node classification and link prediction. We apply learnable output query array in graph classification. The logits layer serves to map the final output to the class dimension.}
\label{tab:input_output}
\end{table}

The GPIO has the capability to perform diverse graph-related tasks such as prediction and classification.
In node classification tasks, the output array obtained after going through the logits layer contains each node's predicted score. We use the cross-entropy loss to learn node embedding. For graph classification tasks, we set a size of output query array to 1-dimension vector. Then we pass the array to the logits layer (e.g., a single linear layer) instead of the read-out layer usually used for global pooling in graph-level tasks. We evaluate the cross-entropy loss with target label. 
For link prediction tasks, we conduct the inner product between each pair of node features which is the output of the final cross attention layer. In the same manner, as GAE \cite{kipf2016variational}, We employ the reconstruction loss for training our model. 
Table \ref{tab:input_output} provides configuration of each array and whether or not the logits layer is used for each task.

\subsection{Dataset}

Table \ref{tab:datasets_graph} describes the dataset statistics for graph-related tasks. For the preprocessing and train/test splits, we follow the benchmark experimental settings for each task and dataset. We follow the same settings with the \cite{Fey:2019wv} for node classification, graph classification and link prediction, respectively.
We provide the dataset as well as the source code as supplementary files. \cf{Note that the graph-related dataset will be automatically downloaded and normalized during preprocessing when running the source code.}

\begin{table}[h!]
\centering
\resizebox{0.65\textwidth}{!}{%
\begin{tabular}{c|ccccccc}
  \toprule
    \textbf{Dataset} & \textbf{Graphs} & \textbf{Nodes} & \textbf{Edges} & \textbf{Features} & \textbf{Classes}  \\
  \midrule
    Cora \cite{cora}    & 1 & 2,708  & 5,278  & 1,433 & 7  \\
    CiteSeer \cite{citeseer} & 1 & 3,327  & 4,552  & 3,703 & 6  \\
    PubMed \cite{pubmed}  & 1 & 19,717 & 44,324 & 500   & 3  \\
  \midrule
    MUTAG   & 188   & 17.93  & 19.79    & 7  & 2 \\
    PROTEINS & 1,113 & 39.06  & 72.82    & 3  & 2 \\
    IMDB-BINARY   & 1,000 & 19.77  & 96.53    & - & 2 \\
    REDDIT-BINARY & 2,00  & 429.63 & 497.754  & - & 2 \\
    COLLAB   & 5,000 & 74.49  & 2,457.22 & - & 3 \\
  \bottomrule
\end{tabular}
}
\caption{Dataset statistics for the graph-related tasks, link prediction, graph classification, and node classification.}
\label{tab:datasets_graph}
\end{table}


\cf{
Cora, CiteSeer, and PubMed are all citation networks. Cora consists of scientific articles classified into seven distinct research areas, each of which is represented by a word vector based on a dictionary of 1,433 words. CiteSeer contains papers classified into six distinct research areas, represented by a word vector using a dictionary of 3,703 words. PubMed includes diabetes-related papers classified into three distinct topics, with each paper represented by a TF-IDF weighted word vector based on a dictionary of 500 words. 
MUTAG is nitroaromatic compounds dataset aimed at predicting their mutagenicity on Salmonella typhimurium. Chemical compounds are represented as graphs, where nodes correspond to (one-hot encoded) atoms and edges represent chemical bonds. 
PROTEINS is a protein dataset classified as enzymes or non-enzymes, where nodes represent amino acids and edges connect those within a distance of less than 6 Angstroms.
IMDB-BINARY is a movie collaboration containing the ego-networks of 1,000 actors and actresses from IMDB. In each graph, nodes represent actors or actresses, and edges connect those who appeared in the same movie. The class of IMDB-BINARY shows reviews as negative or positive.
REDDIT-BINARY is a dataset of Reddit discussion graphs, where nodes represent users and edges indicate replies. The graph is classified as either question/answer-based or discussion-based.
COLLAB is a dataset of scientific collaboration where each graph represents a researcher’s ego network, with nodes as the researcher and collaborators, and edges indicating collaborations. Each graph is labeled based on the researcher’s field: High Energy Physics, Condensed Matter Physics, or Astro Physics.
In our experimental setup, we did not use edge features for any of the graphs, all of which are undirected.}

\cd{Also, we provide dataset statistics for the multimodal text classification.} \cf{The ogbn-products is a graph modeling Amazon's product co-purchasing network, which is undirected and unweighted. Nodes represent products, and edges indicate items bought together. Node features are derived from product descriptions using a bag-of-words approach \cite{chiang2019cluster} with a 100-dimensional feature vector. The goal is to classify products into one of 47 top-level classes.
The ogbn-arxiv dataset is a directed graph modeling citation relationships among Computer Science papers on ARXIV. Nodes represent papers, with edges indicating citations. Each paper is characterized by a 128-dimensional feature vector, computed as the average of word embeddings from its title and abstract \cite{mikolov2013distributed}. For the ogbn-products and ogbn-arxiv datasets, which don't contain edge features, we followed the standard train/validation/test split provided by OGB \cite{hu2020open}.}

\begin{table}[h!]
\centering
\begin{tabular}{lccc}
\hline
Dataset                & \#Nodes & \#Edges   & \#Classes \\ \hline
ogbn-products (subset) & 54,025  & 74,420    & 47        \\
ogbn-arxiv             & 169,343 & 1,166,243 & 40        \\ \hline
\end{tabular}
\caption{\cd{Statistics of OGB datasets \cite{hu2020open}. They contain a large amount of nodes and edges. ogbn-products is subset data as in \cite{he2023harnessing}.}}
\label{tab:ogbn_datasets}
\end{table}

\subsection{\cd{Additional Material}}
\cd{In this section, we provide additional explanatory material to help you understand the paper.}

\subsubsection{\cd{Additional Illustration Material}}

\begin{figure*}[h!] 
    \centerline{\includegraphics[width=1\textwidth]{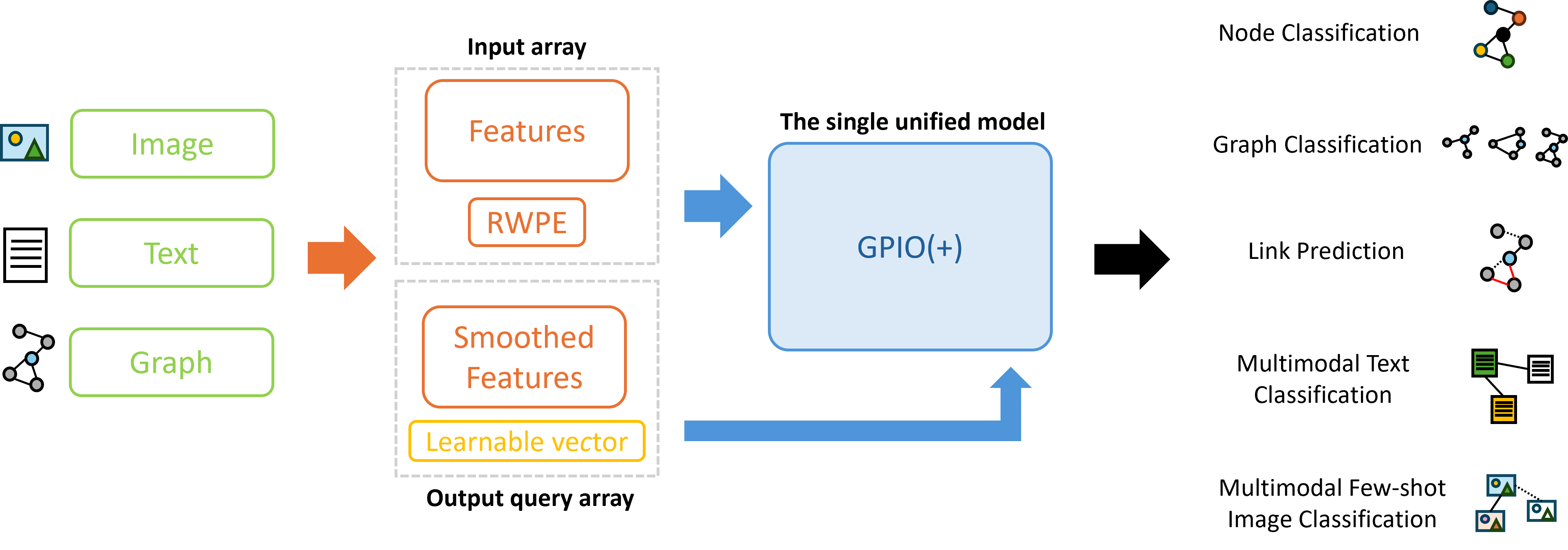}}
    \caption{\cd{GPIO(+), a single architecture, takes multiple modalities such as graph text images as input and output queries (e.g. $RWPE$, Smoothed Features) all in the same way and performs various tasks. The tasks include node classification, graph classification, link prediction, multimodal text classification, and multimodal few-shot image classification.}}
	\label{fig:GPIO_io_all}
\end{figure*}

\cd{
To make it easier to understand at a glance, we have shown in Figure \ref{fig:GPIO_io_all} how a unified model can take different modalities as input and perform various graph-related and multimodal tasks in a unified way.}

\subsubsection{\cd{Space Complexity Analysis}}

\begin{figure*}[h!] 
    \centerline{\includegraphics[width=.9\textwidth]{figures/result_gpu_mem.png}}
    \caption{\cd{GPU memory usage for Barabási-Albert graph densities 0.01, 0.03, and 0.09. GNN 0.09 in the legend means density 0.09 graph. GPIO is invariant to graph density changes and has lower GPU memory usage compared to GNN at high densities.}}
	\label{fig:gpu_mem_compare}
\end{figure*}

\cd{
In Section \textit{Input Array}, GPIO does not use the adjacency matrix $A\in \mathbb{R}^{M \times M}$ in training and evaluation directly, so it offers lower space complexity. 
In practice, most GNNs use a sparse representation and have the space complexity of $O(M \times C + V)$, where $V$ is the number of edges. GPIO, on the other hand, is $O(M \times C)$ since size of the positional encoding $t$ is relatively small. Therefore, the space complexity is all proportional to the number of nodes $M$. However, GPIO is not proportional to $V$. This means that GPIO offer lower space complexity in graphs with a higher density of edges. Figure \ref{fig:gpu_mem_compare} illustrates the GPU memory usage of GNN and GPIO for a given density and number of nodes. We generated graphs with densities of 0.01, 0.03, and 0.09 using the Barabási-Albert model \cite{albert2002statistical}, which captures graph shape similar to real networks. We increased $M$ from about 500 to 8000 and set $C=64$ and $t=32$. Both increase GPU memory usage as $M$ increases, but there is a sharp difference at higher density (e.g. GNN 0.03, GNN 0.09 in the legend). Note taht the GPU memory usage of GPIO is not affected by changes in density, and is only affected by $M$. This result emphasizes that GPIO is efficient in terms of the space complexity required as inputs compared to a typical GNNs when the number of nodes is large, especially for dense graphs. There are many cases that transform the tasks and problems as a graph-structured dataset, such as traveling salesman problems \cite{kool2018attention}; the transformed graph is fully connected.}

\bibliographystyle{elsarticle-num-names} 
\bibliography{cas-refs}

\end{document}